\documentclass[10pt,twocolumn,letterpaper]{article}

\usepackage{iccv}
\usepackage{times}
\usepackage{epsfig}
\usepackage{graphicx}
\usepackage{amsmath}
\usepackage{amssymb}

\usepackage{adjustbox}
\usepackage{times}
\usepackage{microtype}
\usepackage{epsfig}
\usepackage[table,xcdraw]{xcolor}
\usepackage{float}
\usepackage{placeins}
\usepackage{color, colortbl}
\usepackage{stfloats}
\usepackage{enumitem}
\usepackage{tabularx}
\usepackage{xstring}
\usepackage{multirow}
\usepackage{xspace}
\usepackage{url}
\usepackage{subcaption}
\usepackage{xcolor}
\usepackage[hang,flushmargin]{footmisc}
\usepackage{pifont}%
\usepackage{booktabs}
\usepackage{mathtools}
\newcommand{\cmark}{\ding{51}}%
\newcommand{\xmark}{\ding{55}}%

\newcommand{\bY}{\mathbf{Y}}

\newcommand{\bx}{\mathbf{x}}
\newcommand{\by}{\mathbf{y}}

\newcommand{\bM}{\mathbf{M}}

\newcommand{\bD}{\mathbf{D}}

\newcommand{\bP}{\mathbf{P}}

\newcommand{\bff}{\mathbf{f}}
\newcommand{\bF}{\mathbf{F}}

\newcommand{\bQ}{\mathbf{Q}}
\newcommand{\bL}{\mathbf{L}}

\newcommand{\nR}{\mathbb{R}}

\newcommand{\cL}{\mathcal{L}}

\newcommand{\figref}[1]{\Fig~\ref{#1}}
\newcommand{\secref}[1]{Section~\ref{#1}}

\newcommand{\eqnref}[1]{Eq.~\ref{#1}}
\newcommand{\tabref}[1]{Table~\ref{#1}}

\makeatletter
\DeclareRobustCommand\onedot{\futurelet\@let@token\@onedot}
\def\@onedot{\ifx\@let@token.\else.\null\fi\xspace}
\def\eg{e.g\onedot} 
\def\ie{i.e\onedot} 
 
\def\etc{etc\onedot} \def\vs{vs\onedot}

\def\etal{et~al\onedot} 
\def\Fig{Fig\onedot}   
\makeatother

\newcommand{\xdownarrow}[1]{%
  {\left\downarrow\vbox to #1{}\right.\kern-\nulldelimiterspace}
}

\newcommand{\xuparrow}[1]{%
  {\left\uparrow\vbox to #1{}\right.\kern-\nulldelimiterspace}
}

\DeclarePairedDelimiterX{\infdivx}[2]{(}{)}{%
  #1\;\delimsize\|\;#2%
}
\newcommand{\infdiv}{D_{KL}\infdivx}

\DeclareMathAlphabet{\pazocal}{OMS}{zplm}{m}{n}
\newcommand{\unif}{\pazocal{U}}

\newcommand{\boldparagraph}[1]{\vspace{0.15cm}\noindent{\bf #1:} }

\usepackage[pagebackref=true,breaklinks=true,letterpaper=true,colorlinks,bookmarks=false]{hyperref}

 \iccvfinalcopy %

\def\iccvPaperID{1530} %

\ificcvfinal\fi

\begin{document}

\title{RbA: Segmenting Unknown Regions Rejected by All}

\author{
Nazir Nayal$^{1}$ \qquad 
Mısra Yavuz$^{1}$ \qquad
João F. Henriques$^{2}$ \qquad
Fatma Güney$^{1}$ \\
$^1$ KUIS AI Center, Koç University, $^2$ University of Oxford\\
{\tt\small \{nnayal17, myavuz21, fguney\}@ku.edu.tr},
{\tt\small joao@robots.ox.ac.uk}
}

\maketitle
\ificcvfinal\thispagestyle{empty}\fi

\begin{abstract}
Standard semantic segmentation models owe their success to curated datasets with a fixed set of semantic categories,
without contemplating the possibility of identifying unknown objects from novel categories.
Existing methods in outlier detection suffer from a lack of smoothness and objectness in their predictions, due to limitations of the per-pixel classification paradigm. Furthermore, additional training for detecting outliers harms the performance of known classes. 
In this paper, we explore another paradigm with region-level classification to better segment unknown objects. We show that the object queries in mask classification tend to behave like one \vs all classifiers. Based on this finding, we propose a novel outlier scoring function called RbA by defining the event of being an outlier as being rejected by all known classes. 
Our extensive experiments show that mask classification improves the performance of the existing outlier detection methods, and the best results are achieved with the proposed RbA.
We also propose an objective to optimize RbA using minimal outlier supervision. Further fine-tuning with outliers improves the unknown performance, and unlike previous methods, it does not degrade the inlier performance.
\end{abstract}

\section{Introduction}
\label{sec:intro}
We address the problem of semantic segmentation of unknown categories. Detecting novel objects, for example, in front of a self-driving vehicle, is crucial for safety yet very challenging. The distribution of potential objects on the road has a long tail of unknowns such as wild animals, vehicle debris, litter, \etc, manifesting in small quantities on the existing datasets \cite{Yu2020CVPR, Caesar2020CVPR, Cordts2016CVPR}. 
The diversity of unknowns in terms of appearance, size, and location adds to the difficulty. 
In addition to the challenges of data, deep learning has evolved around the closed-set assumption. 
Most existing models for category prediction owe their success to curated datasets with a fixed set of semantic categories.
These models fail in the open-set case by over-confidently assigning the labels of known classes to unknowns~\cite{Hein2019CVPR, Nguyen2015CVPR}.

\begin{figure}[t!]
    \centering
    \includegraphics[width=\linewidth]{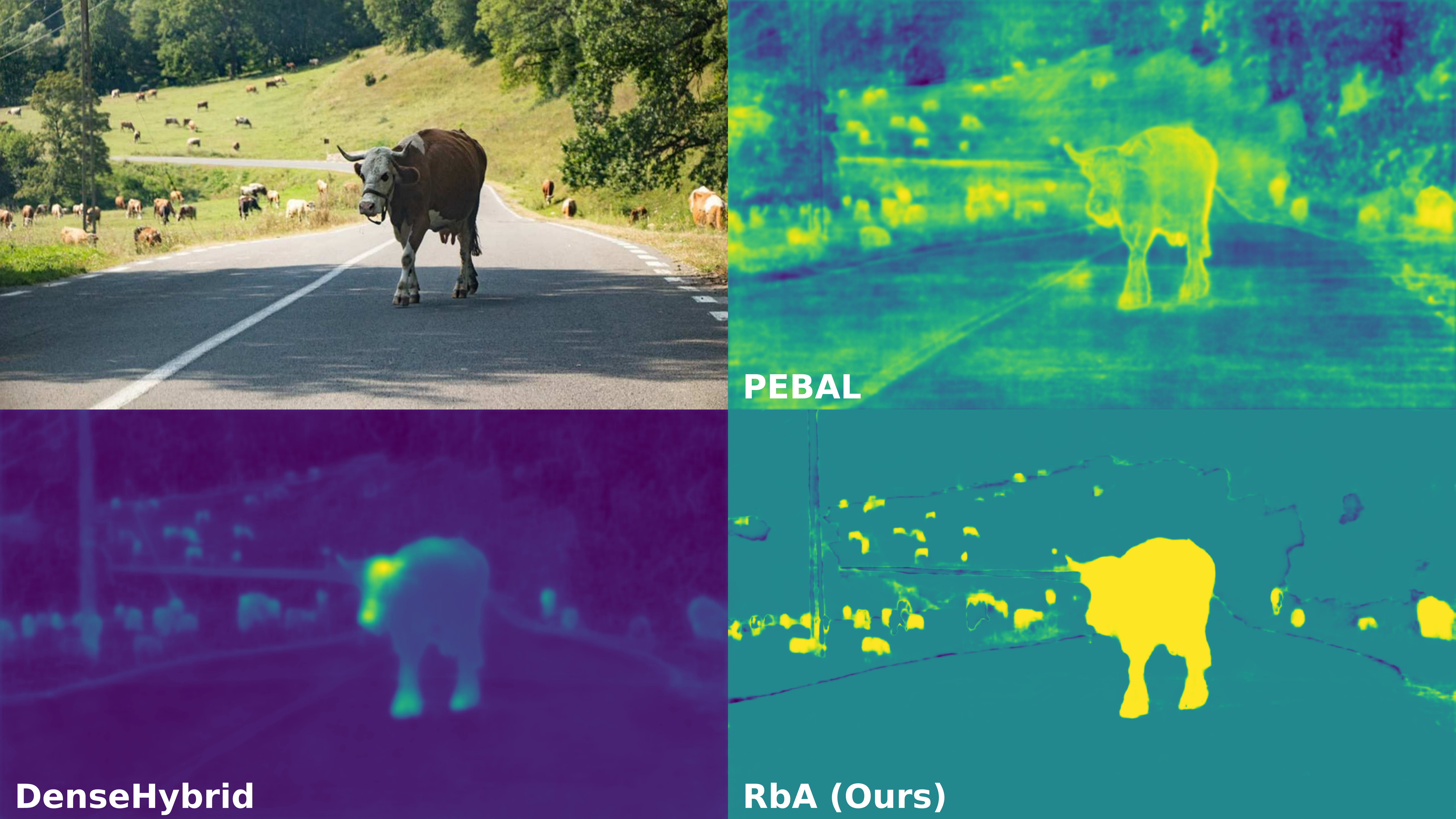}
    \caption{\textbf{Preserving objectness and eliminating noise.} 
    While state-of-the-art methods PEBAL~\cite{Tian2022ECCV} and DenseHybrid~\cite{Grcic2022ECCV} suffer from a lack of smoothness and objectness with high false positive rates, our method RbA clearly segments the unknown objects and reduces false positives by eliminating uncertainty at semantic boundaries and in ambiguous background regions.} %
    \label{fig:qual_teaser}
\end{figure}

The existing approaches to segmenting unknowns 
can be divided into two depending on whether they use supervision for unknown objects or not. In either case, the model has access to known classes during training, \ie inlier or in-distribution, and the goal is to identify the pixels belonging to an unknown class, \ie anomalous, outlier, or out-of-distribution~(OoD). Earlier approaches resort to an ensemble of models~\cite{Lakshminarayanan2017NeurIPS} or Monte Carlo dropout~\cite{Gal2016ICML} which require multiple forward passes, therefore costly in practice. More recent approaches use the maximum class  probability~\cite{Hendrycks2017ICLR} predicted by the model as a measure of its confidence. However, this approach requires the probability predictions to be calibrated, which is not guaranteed~\cite{Szegedy2013ARXIV, Nguyen2015CVPR, Guo2017ICML, Minderer2021NeurIPS, Jiang2018NeurIPS}. In the supervised case, the model can utilize outlier data to learn a discriminative representation, however, outlier data is limited. Typically, another dataset from a different domain is used for this purpose~\cite{Chan2021ICCV}, or outlier objects are artificially added to driving images~\cite{Grcic2022ECCV, Tian2022ECCV}. 

\begin{figure}[t!]
    \centering
    \includegraphics[width=\linewidth] {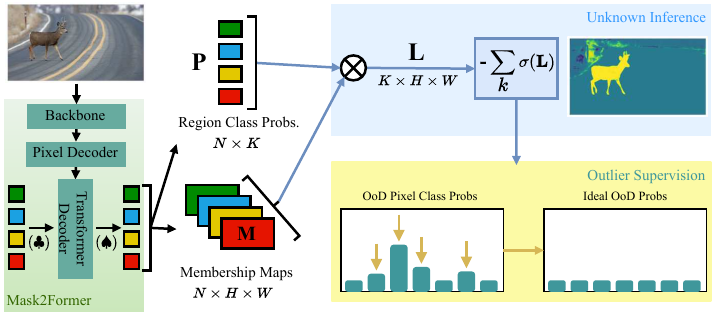}
    \vspace{-0.1in}
    \caption{\textbf{Overview.} This figure provides an illustration of our proposed outlier scoring function RbA and the objective to optimize it as defined in \eqref{eq:in_out_loss_func}. The class logit scores $\bL$ are aggregated as the product of region class probabilities $\bP$ and mask predictions $\bM$ pooled over all regions. We define the RbA as the probability of not being assigned to any of the known classes.
    With the proposed objective, we push the probabilities of known classes down, in the  outlier pixels.
    }
    \vspace{-.1in}
    \label{fig:overview}
\end{figure}

The existing methods in outlier detection suffer from a lack of smoothness and objectness in the OoD predictions as shown in \figref{fig:qual_teaser}. 
This is mainly due to the limitations of the per-pixel classification paradigm that previous OoD methods are built on.
In this paper, we explore another paradigm with region-level classification to better segment objects. To that end, we use mask-classification models, such as Mask2Former \cite{Cheng2022CVPR} that are trained to predict regions and then classify each region rather than individual pixels.
This endows our method with spatial smoothness, learned by region-level supervision. We discover the properties of this family of models which allow better calibration of confidence values. Then, we exploit these properties to boost the performance of the existing OoD methods that rely on predicted class scores such as max logit \cite{Hendrycks2022ICML} and energy-based ones \cite{Grcic2022ECCV, Tian2022ECCV, Liu2020NeurIPS}.

The existing methods also suffer from high false positive rates due to failing to separate the sources of uncertainty, especially on datasets in the wild such as Road Anomaly~\cite{Lis2019ICCV}.
For example, on the boundaries, segmentation models typically predict weak scores for the two inlier classes separated by the boundary, causing these regions to be confused as OoD by score-based methods~\cite{Hendrycks2022ICML}.
Based on exploring the behavior of object queries in mask classification, we find that 
most of the object queries tend to behave like one \vs all classifiers.
Consequently, we propose a novel outlier scoring function based on this one \vs all behavior of object queries.
We define the event of a pixel being an outlier as being rejected by all known classes.
In other words, we define being an outlier as a complementary event whose probability can be expressed in terms of the known class probabilities. We show that this scoring function can eliminate irrelevant sources of uncertainty as in the case of boundaries, resulting in a considerably lower false positive rate on all datasets. %

The state-of-the-art methods in OoD~\cite{Grcic2022ECCV, Tian2022ECCV} utilize outlier data for supervision. While better unknown segmentation can be achieved, it comes at the expense of lower closed-set performance. Unfortunately, this unintended consequence is not desirable since the primary objective of unknown segmentation is to identify unknowns while still accurately recognizing known classes without compromising the inlier performance.

We propose an objective %
to optimize the proposed outlier scoring function
using a limited amount of outlier data. By fine-tuning a very small portion of the model with this objective, our method outperforms the state-of-the-art on challenging datasets with high distribution shifts such as Road Anomaly~\cite{Lis2019ICCV} and SMIYC~\cite{Chan2021ARXIV}. Notably, we achieve this without affecting the closed-set performance.
Our contributions can be summarized as follows:
\begin{itemize}[topsep=0.5em]
    \setlength\itemsep{0em}
    \item We postulate and study the inherent ability of mask classification models to express uncertainty, and use this strength to boost the performance of several existing OoD segmentation methods.
    \item Based on our finding that object queries behave approximately as one \vs all classifiers, we propose a novel outlier scoring function that represents the probability of being an outlier as not being any of the known classes. The proposed scoring function helps to eliminate uncertainty in ambiguous inlier regions such as semantic boundaries.
    \item We propose a loss function that directly optimizes our proposed scoring function using minimal outlier data. The proposed objective exceeds the state-of-the-art by only fine-tuning a very small portion of the model without affecting the closed-set performance.
\end{itemize}

\section{Related Work}
\label{sec:related}
\boldparagraph{Semantic Segmentation Paradigms} Since the success of Fully Convolutional Networks (FCN)~\cite{Shelhamer2015CVPR}, semantic segmentation architectures have revolved around the per-pixel classification paradigm. 
This paradigm has been extensively studied to increase the closed-set performance with various convolution and pooling operations~\cite{Chen2018PAMI, Chen2018ECCV, Dai2017ICCV, Zhao2017CVPR, Yang2018CVPR}, and by aggregating multi-scale contextual information~\cite{Yu2016ICLR, Yuan2020ECCV}. Recent work shifted towards transformer-based architectures~\cite{Xie2021NeurIPS, Strudel2021ICCV, Yuan2021ARXIV, Zheng2021CVPR, Yang2021NeurIPS} and attention mechanisms~\cite{Harley2017ICCV, Li2018BMVC, Zhao2018ECCV, He2019ICCV, Li2019ICCV, Huang2019ICCV, Fu2019CVPR}. 

On the other hand, mask classification has been mainly adopted by instance segmentation and object detection models~\cite{He2017ICCV, Hariharan2014ECCV, Carion2020ECCV} since it allows pixels to belong to multiple proposals and provides the flexibility to detect a variable number of objects in the scene. 
Max-DeepLab~\cite{Wang2021CVPR} employs mask classification for panoptic segmentation but with many auxiliary losses. Although some earlier efforts have been made to apply mask classification to semantic segmentation~\cite{Carreira2012ECCV, Hariharan2014ECCV}, they were quickly outperformed by the per-pixel methods until recently. MaskFormer variants~\cite{Cheng2021NeurIPS, Cheng2022CVPR} apply query-based mask classification and attention to obtain a unified segmentation model which shows competitive performance with specialized semantic and instance segmentation architectures across benchmarks  \cite{Lin2014ECCV, Cordts2016CVPR, Zhou2017CVPR, Neuhold2017CVPR}.

\begin{figure}[t!]
    \centering
    \includegraphics[width=\linewidth] {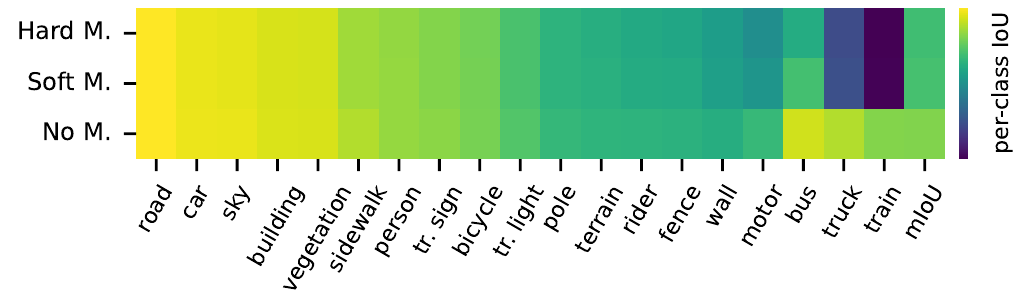}
    \caption{\textbf{Masking object queries.} We show the impact on per-class IoU on Cityscapes \cite{Cordts2016CVPR} when using two types of masking: hard masking without any interactions between the queries (top) and soft masking by allowing interactions in the transformer decoder (middle) compared to the original model without masking (bottom). The constant color in most columns shows that most of the object queries can independently segment their particular classes.
    }
    \label{fig:masking_queries}
\end{figure}

\boldparagraph{Unsupervised Anomaly Segmentation}
Unsupervised methods utilize their knowledge about inlier data to detect anomalies at inference time. Early work measures uncertainty based on the observation that anomaly samples typically result in low-confidence predictions. The uncertainty of a model can be estimated through maximum softmax probabilities~\cite{Hendrycks2017ICLR, Liang2018ICLR}, ensembles \cite{Lakshminarayanan2017NeurIPS}, Bayesian approximation~\cite{Mukhoti2018ARXIV}, Monte Carlo dropout~\cite{Gal2016ICML}, or by learning to estimate its confidence~\cite{Devries2018ARXIV}. However, posterior probabilities of a closed-set model are not necessarily calibrated, leading to overconfident predictions on unseen categories~\cite{Szegedy2013ARXIV, Nguyen2015CVPR, Guo2017ICML, Minderer2021NeurIPS, Jiang2018NeurIPS}. Therefore, follow-up work focuses on making a clear distinction between inliers and outliers by using true class probabilities~\cite{Corbiere2019NeurIPS}, unnormalized logits instead of softmax probabilities~\cite{Hendrycks2022ICML}, standardized class-wise logits~\cite{Jung2021ICCV}, and the distance to learned prototypes of known classes~\cite{Cen2021ICCV}. Overall, unsupervised approaches are typically efficient without any extra training but they are inherently limited to which extent they can separate inliers and outliers due to a lack of supervision with outlier data.

Deep generative models are also used for unsupervised anomaly segmentation. Early methods are primarily based on density estimation~\cite{Lee2018NeurIPS, Ren2019NeurIPS} while subsequent works focus on reconstruction.
Several works rely on the predicted segmentation maps to resynthesize~\cite{Lis2019ICCV, Xia2020ECCV, Haldimann2019ARXIV, Vojir2021ICCV} or inpaint the inputs~\cite{Lis2020ARXIV} and measure discrepancy with comparison networks. Others apply localized adversarial attacks~\cite{Besnier2021ICCV}, synthesize negatives using normalizing flows~\cite{Grcic2021ARXIV}, or combine Gaussian mixture models with discriminative representation learning~\cite{Liang2022ARXIV}. 
Generative methods are typically impractical for real-time safety-critical applications due to high computational costs and long inference times. Additional comparison modules and the change in input distributions require extra training. 
Moreover, synthesized unknowns often do not generalize well to real anomalies~\cite{Grcic2021ARXIV}. Several works~\cite{Nalisnick2018ICLR, Serr2020ARXIV, Zhang2021ICML} show that generative models tend to estimate high likelihoods on out-of-distribution samples. 

\boldparagraph{Anomaly Segmentation with Outlier Supervision}
Out-of-distribution data can be used to regularize the model's feature space by learning a representation of unknowns.
With the increase in the availability of wide-range datasets, initial approaches utilize generic datasets such as ImageNet \cite{Russakovsky2015IJCV} %
for OoD. Given data, OoD detection can be simply treated as binary classification~\cite{Bevandic2018ARXIV, Bevandic2019GCPR}. Outlier data can also be used to estimate the distributional uncertainty of OoD samples~\cite{Malinin2018NeurIPS} or to fine-tune parametrized OoD detectors~\cite{Hendrycks2019ICLR}.
The energy score has been proposed as a better alternative to softmax in terms of separation~\cite{Liu2020NeurIPS}. SynBoost~\cite{Biase2021CVPR} is a supervised image resynthesis method that treats void regions as anomalies to obtain an uncertainty signal.

Recent work uses a subset of COCO \cite{Lin2014ECCV} or ADE20K \cite{Zhou2017CVPR}, either as entire images~\cite{Chan2021ICCV} or after cut-and-paste into the inlier scenes~\cite{Tian2022ECCV, Grcic2022ECCV}. Meta-OoD~\cite{Chan2021ICCV} maximizes the entropy on outliers, whereas PEBAL~\cite{Tian2022ECCV} learns adaptive energy-based penalties by abstention learning. Combining likelihood and posterior evaluation, DenseHybrid~\cite{Grcic2022ECCV} achieves state-of-the-art results. However, for each benchmark, different models are fine-tuned using multiple datasets~\cite{Zhou2017CVPR, Neuhold2017CVPR, Zendel2018ECCV} with high distribution shifts, resulting in a higher degree of supervision and variety. Our model, on the other hand, can achieve better performance across benchmarks by using the same model and only a small subset of COCO~\cite{Lin2014ECCV} for fine-tuning.

\section{Methodology}
\label{sec:method}
In this work, we address the limitations of the existing OoD methods by using mask classification.
We first perform an analysis of the mask classification models. Then, based on our analysis, we propose a novel scoring function to exploit the implicit one \vs all behavior in these models. We mathematically define the probability of being an outlier probability as the ``none of the above" option for the model. Finally, we propose a training objective to optimize our proposed scoring function with minimal outlier data.

\subsection{Mask Classification}
We build our method on top of the Mask2Former architecture \cite{Cheng2022CVPR}, which is an improved version of the initial MaskFormer~\cite{Cheng2021NeurIPS}.
We give only a brief overview to make the discussion self-contained; please refer to Cheng \etal \cite{Cheng2022CVPR} for details.
Mask2Former consists of three main parts: the backbone, the pixel decoder, and the transformer decoder. The backbone processes the input image $\bx \in \mathbb{R}^{3 \times H \times W}$ %
to extract features at multiple scales. Then, the pixel decoder further processes the multi-scale features to produce high-resolution per-pixel features $\bF(\bx) \in \nR^{C_{p} \times H \times W}$.
The transformer decoder takes the resulting multi-scale features $\{\bff_i\}_{i=1}^{D}$, where $D$ is number of scales, as well as $N$ learnable object queries $\bQ \in \nR^{N \times C_{q}}$, where $C_{p}$ and $C_{q}$ denote the embedding dimensions. 
At each layer of the transformer decoder, object queries are refined by interacting with each other and with one of the scales $\bff_i$ in a round-robin order. 

The refined object queries are first processed with a 3-layer MLP, resulting in $\bQ_{p} \in \nR^{N \times C_{p}}$ to predict $N$ regions. The binary masks for all regions are obtained by multiplying $\bQ_{p}$ with pixel features $\bF$ and applying a sigmoid $\sigma$ to the result:
\begin{equation}
    \bM(\bx) = \sigma(\bQ_p~\bF(\bx))
\end{equation}
$\bM(\bx) \in \nR^{N \times H \times W}$ represents the membership score of each pixel belonging to a region.
In parallel, refined object queries are fed to a linear layer followed by softmax to produce posterior class probabilities $\bP(\bx) \in [0, 1]^{N \times K}$ of $K$ classes.

In contrast to per-pixel semantic segmentation, the ground truth masks are partitioned into multiple binary masks such that each mask contains all the pixels that belong to a class. Then, bipartite matching is used to match every ground truth mask to an object query using region prediction and classification losses as the cost. For region prediction, 
a weighted combination of dice loss~\cite{Milletari2016THREEDV} and binary cross-entropy is applied to the binary mask predictions. For classification, %
cross-entropy loss is used.
In inference, the class scores or logits $\bL(\bx) \in \nR^{K \times H \times W}$ are calculated as the product of mask predictions with class predictions by broadcasting the class prediction to all the pixels within the region:
\begin{equation}
    \bL(\bx) = \sum_{n=1}^{N} \bP_n(\bx) \bM_n(\bx)
    \label{eq:vote}
\end{equation}

\subsection{Independence of Object Queries}
\label{sec:one_vs_all}
The logit term $\bL$ as defined in \eqnref{eq:vote} has a deeper interpretation because of its structure. In essence, $\bL$ aggregates weighted votes over all object queries to decide whether the pixel belongs to a certain class. During training, the ground truth binary map of each class is matched to an object query using bipartite matching. Therefore, we find that object queries specialize in predicting a specific class after convergence. We empirically verify this behavior on another driving dataset (after training on Cityscapes), the validation set of BDD100K~\cite{Yu2020CVPR}. We identify which class each object query specializes in by counting how many times it predicts a certain class with high confidence, \eg greater than $98\%$, see Supplementary for visualization of this specialized behavior.

After identifying which object query predicts which class, we test their independence, \ie the ability of each object query to predict its class without relying on other object queries. 
To evaluate the predictions of class $k$, we mask out all but its specialized query.
We do this in one of two ways: 1) before the transformer decoder ($\clubsuit$ in \figref{fig:overview}), where each object query only interacts with the image features and not with each other (\textit{hard masking}), or 2) after the transformer decoder ($\spadesuit$ in \figref{fig:overview}), allowing queries to interact with each other weakly (\textit{soft masking}). In both cases, only the specialized query is used to predict the mask and class.

\figref{fig:masking_queries} shows per-class IoU scores on Cityscapes \cite{Cordts2016CVPR} using both strategies compared to the original model without any masking. We observe that the performance in most of the classes is not affected compared to the original model. The drop in performance occurs only in rare classes, such as train, truck, or bus, indicating that their object queries rely on other queries in prediction, which explains the slightly better performance of soft masking than hard masking. This behavior of object queries %
resembles multiple independent binary classifiers, implicitly embedded in a single model. Note that this is only for analysis purposes and not part of the proposed method.

\begin{figure*}[t!]
    \centering
    \includegraphics[width=\linewidth]{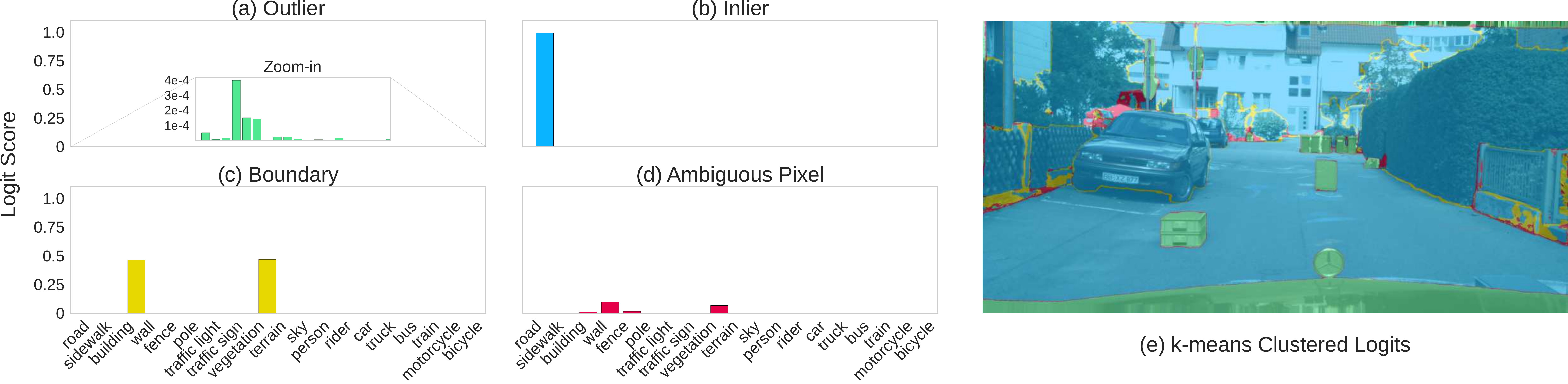}
    \caption{\textbf{Categorizing the behavior of logits.} Outlier pixels receive extremely low votes from object queries \textcolor{red}{(a)}. Inlier pixels receive a high vote from a single object query \textcolor{red}{(b)}. Boundary pixels separating two inlier classes receive moderate votes from both object queries \textcolor{red}{(c)}. Ambiguous regions receive weak votes from multiple object queries \textcolor{red}{(d)}. Clustering clearly outlines these four behaviors of logits \textcolor{red}{(e)}. Pixels in \textcolor{red}{(e)} are color-coded with the same colors of the respective histograms \textcolor{red}{(a-d)}.}
    \label{fig:logits}
\end{figure*}

\subsection{Rejected by All (RbA) Scoring Function}
Inspired by the independent behavior of object queries, we propose to model the prediction of each class as an independent binary classification problem.
We consider it as $K$ one \vs all classifiers where the predicted score for each class is independently modeled as follows:
\begin{equation}
    p(\by = k|\bx) = \sigma(\mathbf{L}_k(\bx))
\end{equation}
where $\by \in \mathcal{K}^{H \times W}$ is a random variable representing the predicted class label over a predefined set of known classes $\mathcal{K} = \{1, \dots, K\}$
and $\sigma$ is a normalization function applied to per class logits to map them to a probability, \ie a value between $0$ and $1$. Based on this definition, we assume that the latent space is partitioned into $K+1$ mutually exclusive and exhaustive regions, such that the label $K+1$ represents the region where the outliers reside, rejected by all other known classes. By assuming the mutual exclusiveness of per-class probabilities for a given input, we can define the probability map of an input $\bx$ being an outlier as follows:
\begin{equation}
    \begin{split}
        p(\by = K+1|\bx) &= 1 - p\left(\cup_{k=1}^{K} \by = k \vert \bx\right) \\
                         &= 1 - \sum_{k=1}^{K}p(\by = k \vert \bx) \\
                         &= 1 - \sum_{k=1}^{K}\sigma(\mathbf{L}_k(\bx))
    \end{split}
\end{equation}
Dropping the constant $1$ (which does not affect the optimization), we define our outlier scoring function RbA:
\begin{equation}
    \mathrm{RbA}(\bx) = -\sum_{k=1}^{K}{\sigma(\bL_k(\bx))}
    \label{eq:nls}
\end{equation}
We choose $\sigma$ to be the \texttt{tanh} function to map $\bL_k > 0$ more uniformly to the range $[0, 1]$.

\subsection{Fine-tuning with Minimal Outlier Supervision}
\label{sec:outlier_supervision}
We propose to regularize our scoring function with supervision from a small amount of synthetically created outlier data. Our goal is to improve the OoD segmentation while preserving the closed-set performance. Without retraining the entire model, we only fine-tune the mask prediction MLP and classification layer after the transformer decoder (see Supplementary), which constitutes only $0.21\%$ of the total model parameters. For OoD data, we use a modified version of Anomaly Mix proposed in \cite{Tian2022ECCV}, where objects from COCO dataset~\cite{Lin2014ECCV} are randomly cut and pasted on Cityscapes images~\cite{Cordts2016CVPR}. We regularize the scores by \emph{maximizing RbA for outlier pixels}, with a squared hinge loss. This is also equivalent to suppressing high-confidence probabilities of known classes for outlier pixels as shown in \figref{fig:overview}. The loss is formally defined as follows:
\begin{eqnarray}
    \label{eq:in_out_loss_func}
    \cL_{\mathrm{RbA}} &=& \sum_{\bx \in \Omega_{out}} \left(\max(0, \alpha - \text{RbA}(\bx))\right)^2 \\
     &=& \sum_{\bx \in \Omega_{out}} \max\left(0, \alpha + \sum_{k=1}^{K}{\sigma(\bL_k(\bx)}) \right)^2 \nonumber%
\end{eqnarray}
where 
$\Omega_{out}$ is the set of outlier pixels. We experimentally set the hyper-parameter $\alpha$ to 5 but we found that any $\alpha > 0$ works well in practice. See Supplementary for an ablation.

\subsection{Analyzing RbA}
The term $\bL$ in \eqnref{eq:vote} aggregates the independent decisions of object queries about whether a pixel belongs to a certain class. Based on this behavior, we can identify several distinct modes of $\bL$. We cluster the logits over classes at each pixel, \ie $K$-dimensional vector, using k-means to characterize the modes, visualized in \figref{fig:logits}\textcolor{red}{e}. For an inlier pixel, only a single object query votes for it with high confidence (\figref{fig:logits}\textcolor{red}{b}), whereas true outlier pixels do not receive any votes from any object query (\figref{fig:logits}\textcolor{red}{a}). 
These two modes, %
especially the outliers in \figref{fig:logits}\textcolor{red}{a}, due to the one \vs all behavior, reduces the overconfidence issue in the existing outlier scoring functions used in max logit~\cite{Hendrycks2022ICML} and energy-based methods~\cite{Grcic2022ECCV, Tian2022ECCV, Liu2020NeurIPS}, therefore improve their results (\tabref{tab:method}).

\begin{figure*}[t!]
    \centering
    \includegraphics[width=\linewidth]{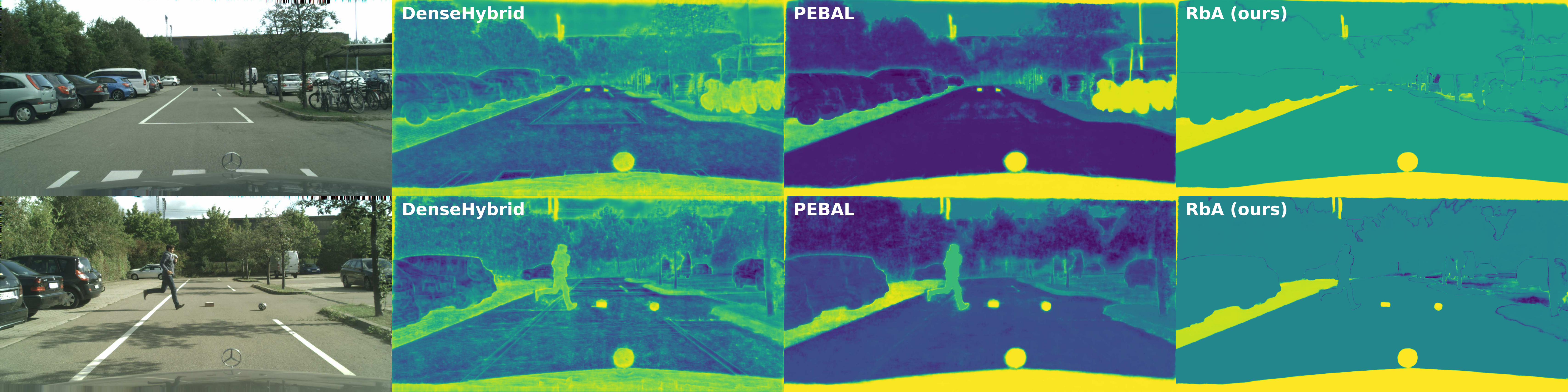}
    \caption{\textbf{Visual comparison to the state-of-the-art}. We show visualizations of outlier score maps predicted by our method, RbA compared to the ones predicted by state-of-the-art methods PEBAL~\cite{Tian2022ECCV} and DenseHybrid~\cite{Grcic2022ECCV} trained using the same architecture as the RbA for a fair comparison. The other two methods falsely identify the inlier classes such as person and bike, which are correctly ignored by the proposed RbA. It is noteworthy that RbA also eliminates the false positives in the background region, especially at the boundaries separating inliers and better preserves the smoothness of the outlier map compared to other methods despite being also trained with mask classification.}
    \label{fig:qual_ml_vs_nls}
\end{figure*}

However, there are pixels that disrupt the separability between the inliers and the outliers which max logit and energy-based methods fail to capture. For example, pixels on a boundary between two inlier classes (\figref{fig:logits}\textcolor{red}{c}) or ambiguous background pixels (\figref{fig:logits}\textcolor{red}{d}) end up with a higher anomaly score 
than the inliers, causing them to be mistaken as an outlier. 
Boundary and ambiguous regions are commonly characterized by having more than one weak vote from object queries. Since RbA aggregates votes from all classes, summing these weak votes results in a lower outlier score and hence reduces the false positive rate.
\figref{fig:qual_ml_vs_nls} highlights the differences between the anomaly maps predicted by RbA and the state-of-the-art methods, also trained using Mask2Former. Note that RbA assigns low outlier scores at boundaries separating known classes.

\section{Experiments}
\label{sec:experiments}
\begin{table*}[t!]
    \centering
    \adjustbox{max width=\textwidth}{%
    \begin{tabular}{l c c  c c c c c c c c c c } 
    \toprule
    \multirow{2}{*}{Method} & \multirow{1}{*}{OoD} & \multirow{1}{*}{Extra}  & \multicolumn{5}{c}{Anomaly Track}  & \multicolumn{5}{c}{Obstacle Track} \\      \cmidrule(r){4-5} \cmidrule(lr){6-8} \cmidrule(lr){9-10} \cmidrule(l){11-13}
           &  Data & Net. & AP $\uparrow$ & FPR $\downarrow$ & sIoU gt $\uparrow$ & PPV $\uparrow$ & mean F1 $\uparrow$ & AP $\uparrow$  & FPR $\downarrow$ & sIoU gt $\uparrow$ & PPV $\uparrow$ & mean F1 $\uparrow$ \\
    \midrule
     Emb. Density\cite{Blum2021IJCV}                & \xmark & \xmark & 37.5 & 70.8 & 33.9 & 20.5 & 7.9 & 0.8 & 46.4 & 35.6 & 2.9 & 2.3\\
     JSRNet\cite{Vojir2021ICCV}                    & \xmark & \cmark & 33.6 & 43.9 & 20.2 & 29.3 & 13.7 & 28.1 & 28.9 & 18.6 & 24.5 & 11.0 \\
     Road Inpain.\cite{Lis2020ARXIV}               & \xmark & \cmark & -     & -     & -     & -     & -  & 54.1 & 47.1 & \textbf{57.6} & 39.5 & \underline{36.0}  \\
     Image Resyn.\cite{Lis2019ICCV}                & \xmark & \cmark & 52.3 & \underline{25.9} & 39.7 & 11.0 & 12.5 & 37.7 & 4.7 & 16.6 & 20.5 & 8.4 \\ 
     ObsNet\cite{Besnier2021ICCV}                  & \xmark & \cmark & \underline{75.4} & 26.7 & \underline{44.2} & \textbf{52.6} & \textbf{45.1} & - & - & - & - & -  \\
     NFlowJS\cite{Grcic2021ARXIV}                  & \xmark & \cmark & 56.9 & 34.7 & 36.9 & 18.0 & 14.9 & \underline{85.6} & \textbf{0.4} & 45.5 & \underline{49.5} & \textbf{50.4}  \\
     RbA (Ours)                                     & \xmark & \xmark & \textbf{86.1} & \textbf{15.9} & \textbf{56.3} & \underline{41.4} & \underline{42.0} & \textbf{87.8} & \underline{3.3} & \underline{47.4} & \textbf{56.2} & \textbf{50.4}  \\ %
     \midrule
     Max. Entropy\cite{Li2019ICCV}                  & \cmark & \xmark & \underline{85.5} & 15.0 & 49.2 & \underline{39.5} & 28.7 & 85.1 & 0.8 & \underline{47.9} & \textbf{62.6} & 48.5 \\ 
     DenseHybrid\cite{Grcic2022ECCV}               & \cmark & \xmark & 78.0 & \textbf{9.8} & \underline{54.2} & 24.1 & \underline{31.1} & \underline{87.1} & \textbf{0.2} & 45.7 & 50.1 & \underline{50.7}  \\ 
     PEBAL\cite{Tian2022ECCV}                      & \cmark & \xmark & 49.1 & 40.8 & 38.9 & 27.2 & 14.5 & 5.0 & 12.7 & 29.9 & 7.6 & 5.5 \\
     SynBoost\cite{Biase2021CVPR}                  & \cmark & \cmark & 56.4 & 61.9 & 34.7 & 17.8 & 10.0 & 71.3 & 3.2  & 44.3 & 41.8 & 37.6 \\
     RbA (Ours)        & \cmark & \xmark & \textbf{90.9} & \underline{11.6} & \textbf{55.7} & \textbf{52.1}  & \textbf{46.8} & \textbf{91.8} & \underline{0.5} & \textbf{58.4} & \underline{58.8} & \textbf{60.9 } \\ 
     \bottomrule
    \end{tabular}}
    \caption{\textbf{Results on the SMIYC benchmark.} We report results on both the anomaly and the obstacle track. Both tracks cover a wide variety of scenarios and unknown objects. We report both pixel-level (AP, FPR@95) and component-level metrics (sIoU, PPV, mean F1). We show the results with (lower part) and without (upper part) outlier supervision with the best in bold and the second best underlined for each.} %
    \vspace{-.05in}
    \label{tab:smiyc_merged}
\end{table*}

\subsection{Datasets}
We train the model on Cityscapes \cite{Cordts2016CVPR}, which consists of 2975 training and 500 validation images. It contains 19 classes which are considered as inliers in anomaly segmentation benchmarks. The classes in the dataset can be seen in \figref{fig:masking_queries}. For evaluation, we consider multiple datasets. First, Segment Me If You Can~(SMIYC) benchmark \cite{Chan2021ARXIV} with two datasets: anomaly track and obstacle track. The anomaly track has 100 images that contain unknown objects of various sizes in diverse environments. The obstacle track contains 412 images with typically small unknown objects on the road, 85 of which are taken at night and in adverse weather conditions. Both datasets are characterized by a high domain shift compared to Cityscapes, making them particularly challenging. 
Road Anomaly~\cite{Lis2019ICCV} is an earlier and smaller version of SMIYC. It consists of 60 images with diverse objects in diverse environments. 
We also report results on the Fishyscapes Lost\&Found \cite{Blum2021IJCV}, which has 100 validation and 275 test images. 
The domain of this dataset is similar to that of Cityscapes, and the anomalous objects are mostly small and less diverse compared to other datasets. 

\subsection{Experimental Setup}
\boldparagraph{Implementation Details} 
We follow the setup of \cite{Cheng2022CVPR} for closed-set training on Cityscapes. We use the Swin-B~\cite{Liu2021ICCV} architecture as the backbone. Differently, we use only one decoder layer in the transformer decoder instead of nine (see Supplementary).  %
For outlier supervision, we fine-tune the mask prediction MLP and the classification layer for 2K iterations with a batch size of 16 using the standard loss functions used in \cite{Cheng2022CVPR} in addition to the RbA loss defined in \eqnref{eq:in_out_loss_func}. 
Previous work~\cite{Tian2022ECCV} samples 300 new images every epoch out of 40K COCO images with objects different than Cityscapes inliers. %
Differently, we sample 300 images only at the beginning and fix them, then at each iteration, an image is randomly chosen and pasted on inlier images with probability $p_{out}$. We experimentally set $p_{out}$ to $0.1$.

\boldparagraph{Evaluation Metrics} 
For comparison to previous methods on the Road Anomaly and the Fishyscapes, we report Average Precision~(AP), Area under ROC Curve~(AuROC), and False Positive rate at the threshold of $95\%$ True Positive Rate~(FPR@95). On SMIYC, the public benchmark reports AP and FPR@95 for per-pixel metrics as well as component-level metrics that are designed to measure the statistics of detected objects \cite{Chan2021ARXIV}. Specifically, the proposed metrics aim at quantifying true positives (TP), false negatives (FN), and false positives (FP) of detected unknown objects. Please see the benchmark paper~\cite{Chan2021ARXIV} for more details on these metrics.

\begin{table*}[t!]
\footnotesize
    \centering
    \begin{tabular}{l c c c c c c c c } 
     \toprule
     \multirow{2}{*}{Method} & \multirow{1}{*}{OoD} & \multirow{1}{*}{Extra } & \multicolumn{3}{c}{Road Anomaly} & \multicolumn{3}{c}{FS LaF}  \\
        \cmidrule(r){4-6} \cmidrule(l){7-9}
        &  Data & Net. &AUC $\uparrow$ & AP $\uparrow$ & FPR $\downarrow$ & AUC $\uparrow$ & AP $\uparrow$ & FPR $\downarrow$ \\
         \midrule
         MSP (R101) \cite{Hendrycks2017ICLR}        & \xmark & \xmark & 73.76 & 20.59 & 68.44 & 86.99 & 6.02  & 45.63 \\
         Entropy (R101) \cite{Hendrycks2017ICLR}    & \xmark & \xmark & 75.12 & 22.38 & 68.15 & 88.32 & 13.91 & 44.85 \\
         Mahalanobis \cite{Lee2018NeurIPS}          & \xmark & \xmark & 76.73 & 22.85 & 59.20 & 92.51 & 27.83 & 30.17 \\
         SML \cite{Jung2021ICCV}                    & \xmark & \xmark & 81.96  &25.82 & 49.74 & \underline{96.88} & 36.55 & 14.53 \\ 
         GMMSeg (SF) \cite{Liang2022ARXIV}          & \xmark & \xmark & \underline{89.37} & \underline{57.65} & \underline{44.34} & \textbf{97.83} & \underline{50.03} & \underline{12.55} \\
         SynthCP \cite{Xia2020ECCV}                 & \xmark & \cmark & 76.08 & 24.86 & 64.69 & 88.34 & 6.54  & 45.95 \\
         RbA (Ours)                                 & \xmark & \xmark & \textbf{95.60} & \textbf{78.45}  & \textbf{11.83} & 96.43 & \textbf{60.96} & \textbf{10.63} \\
         \midrule
         Maximized Entropy \cite{Chan2021ICCV}      & \cmark & \xmark & -     & -     & -     & 93.06 & 41.31 & 37.69 \\
         PEBAL \cite{Tian2022ECCV}                  & \cmark & \xmark & \underline{88.85} & \underline{44.41} & \underline{37.98} & \underline{98.52} & \underline{64.43} & \underline{6.56}  \\ 
         SynBoost (WRN38) \cite{Biase2021CVPR}      & \cmark & \cmark & 81.91 & 38.21 & 64.75 & 96.21 & 60.58 & 31.02 \\
        RbA (Ours)                        & \cmark & \xmark & \textbf{97.99} & \textbf{85.42}	& \textbf{6.92} & \textbf{98.62} & \textbf{70.81} & \textbf{6.30} \\
         \bottomrule   
        \end{tabular}
        \caption{\textbf{Results on Road Anomaly and Fishyscapes LaF.} %
        We show the results with (lower part) and without (upper part) outlier supervision with the best in bold and the second best underlined for each. We report the results of RbA both with and without outlier supervision. 
        Our method RbA notably improves the results in all metrics on both datasets.
        } %
        \vspace{-.05in}
        \label{tab:ra_fs_val}
\end{table*}

\subsection{Quantitative Results}
\subsubsection{Segment Me If You Can Benchmark}
\tabref{tab:smiyc_merged} shows the results on anomaly and obstacle tracks of the public SMIYC benchmark. Without outlier supervision, RbA outperforms all the models, including those trained with outlier supervision, in AP while maintaining a competitive FPR@95. In terms of component metrics, the gains with RbA are more pronounced, which is due to an improved ability to characterize objectness, compared to the previous work. With outlier supervision, the performance gap improves with respect to the previous best method %
consistently across both tracks: +5.4\%  and +4.7\% in AP and +1.7\% and +10.2\% in mean F1 for anomaly and obstacle tracks respectively.
DenseHybrid~\cite{Grcic2022ECCV} achieves a slightly better FPR@95 on the anomaly and obstacle tracks, but RbA achieves significantly better AP, +12.9\% and +4.7\% respectively, and better performance in all component-level metrics. 
ObsNet \cite{Besnier2021ICCV} has impressive performance at the component-level, however, not at the pixel-level. RbA consistently performs well across both tracks in both pixel and component-level metrics.

SMIYC is characterized by high domain shift and diversity of objects in terms of size and appearance, making it particularly challenging. While some methods, like DenseHybrid~\cite{Grcic2022ECCV}, rely on highly diverse data when fine-tuning, RbA with mask classification shows that outlier supervision is not necessary to perform well under domain shift, thereby surpassing the limitations of the existing methods.

\subsubsection{Road Anomaly \& Fishyscapes LaF}
\tabref{tab:ra_fs_val} shows the results on the Road Anomaly~\cite{Lis2019ICCV} and the Fishyscapes Lost and Found (LaF) validation set~\cite{Blum2021IJCV}. 
Without outlier supervision, RbA improves the state-of-the-art significantly in almost all metrics on both datasets, even outperforming methods with outlier supervision in some metrics. With minimal supervision from a limited number of outlier objects, we obtain significant performance gains without hurting the closed-set performance (\tabref{tab:method}).

\subsection{Ablation Study}
\label{sec:ablations}

We ablate our contributions to justify our decision choices with the scoring function, loss function, and backbone. First, we show that mask classification improves the performance of the existing methods in OoD, but RbA better utilizes its potential. We then report the performance of the squared hinge loss compared to alternative loss functions. Lastly, we experiment with different backbones and show that optimizing for RbA improves the results with different backbones.

\begin{table}[b!]
\footnotesize
\centering
\adjustbox{max width=\columnwidth}
\centering
    \begin{tabular}{l c c c  c c} 
     \toprule
     \multirow{2}{*}{Method} & \multirow{2}{*}{mIoU $\uparrow$} & \multicolumn{2}{c}{Road Anomaly} & \multicolumn{2}{c}{FS LaF}  \\
         \cmidrule(r){3-4} \cmidrule(l){5-6}
        & & AP $\uparrow$ & FPR $\downarrow$ & AP $\uparrow$ & FPR $\downarrow$ \\
     \midrule
     Max Logit~\cite{Hendrycks2022ICML}& \textbf{82.25} & 77.31 & 16.90 & 58.52 & 22.14 \\
     PEBAL~\cite{Tian2022ECCV} & 75.32 & \underline{79.01} & \underline{7.21} & \underline{62.67} & 25.60 \\
     DH~\cite{Grcic2022ECCV} & 80.27 & 78.57 & 12.28 & 36.94 & \underline{21.12} \\
     RbA (Ours) & \underline{82.20} & \textbf{85.42} & \textbf{6.92} & \textbf{70.81} & \textbf{6.30} \\
     \bottomrule
    \end{tabular}
    \caption{\textbf{Other methods with Mask2Former.} We show the performance of the state-of-the-art methods with Mask2Former. %
    Our method RbA achieves the best results in all metrics with a clear margin, without affecting the closed-set performance, unlike previous methods. The mIoU before fine-tuning is shown in the first row of the table. The rest of the models are fine-tuned from the same checkpoint.}
    \label{tab:method}
\end{table}

\boldparagraph{Other Methods with Mask Classification} To clearly demonstrate the effectiveness of our method and decouple it from the gains obtained by the Mask2Former, we report the results of other SOTA methods using Mask2Former, including PEBAL~\cite{Tian2022ECCV}, DenseHybrid~\cite{Grcic2022ECCV}, and Max Logit~\cite{Hendrycks2022ICML} in \tabref{tab:method}. 
The existing OoD methods perform well with Mask2Former, for example, the performance of PEBAL significantly improves compared to the official results reported in \tabref{tab:ra_fs_val}.
As discussed in \secref{sec:one_vs_all}, the improvement comes from reducing the overconfidence issue owing to the independent behavior of object queries. Our method, RbA, performs better than the other methods in all metrics. More importantly, we achieve this performance in OoD without affecting the closed-set performance, unlike the other methods such as PEBAL causing a significant drop in mIoU. This experiment shows that we can better utilize the properties of mask classification with RbA.

\begin{table}[h!]
\footnotesize
\centering
    \begin{tabular}{l c c c c c} 
     \toprule
     \multirow{2}{*}{Method} & \multicolumn{2}{c}{Road Anomaly} & \phantom{ab} & \multicolumn{2}{c}{FS LaF}  \\
     \cmidrule{2-3} \cmidrule{5-6}
        & AP $\uparrow$ & FPR $\downarrow$ & & AP $\uparrow$ & FPR $\downarrow$ \\
     \midrule
     
     KL Div.     & 79.91             & 11.33       &      & 63.58             & 8.78 \\
     MSE         & 80.71             & 15.79       &      & \underline{69.14} & 22.06 \\
     L1          & \underline{80.94} & 15.75       &      & 67.19             & 20.44 \\
     BCE         & 80.66	            & \underline{10.29} & & 64.90             & \underline{6.89} \\
     RbA (Ours)  & \textbf{85.42}	& \textbf{6.92}    & & \textbf{70.81}    & \textbf{6.30}\\
     \bottomrule
    \end{tabular}
    \caption{\textbf{Ablation study on alternative loss functions.} We compare our loss function based on the squared hinge loss to other commonly used loss functions. The results show that our method with squared hinge loss (RbA) performs the best in terms of OoD segmentation.} %
    \label{tab:abl_loss}
\end{table}

\boldparagraph{Alternative Loss Functions} We verify our choice of loss function which is a squared hinge loss by optimizing our method with other commonly used loss functions. %
As can be seen in \tabref{tab:abl_loss}, squared hinge loss outperforms other loss functions. Mean Squared Error (MSE) and L1 result in a higher false positive rate. We define the OoD as a binary classification problem and optimize it with BCE by using the outlier score given by the RbA as the positive class logit. While it improves the FPR compared to MSE and L1, it performs worse than the squared hinge loss in all metrics. Using KL Divergence, we minimize the distance between class probabilities of outlier pixels from a fixed distribution with maximum entropy. It performs comparably in FPR but poorly in AP, especially on the Fishyscapes LaF. Detailed formulations can be found in Supplementary.

\begin{table}[h]
\footnotesize
\centering
\begin{tabular}{l c c c c c} 
     \toprule
     \multirow{2}{*}{Backbone}  & \multicolumn{2}{c}{Road Anomaly}  &  & \multicolumn{2}{c}{FS LaF}  \\
     \cmidrule{2-3} \cmidrule{5-6}
                                & AP $\uparrow$ & FPR $\downarrow$ & & AP $\uparrow$ & FPR $\downarrow$ \\
    
     \midrule
      R101~\cite{He2016CVPR}  & 38.1 / 61.9 & 82.7 / 37.2 & & 30.1 / 47.1 & 26.3 / 12.7 \\
      WR38~\cite{Zagoruyko2016BMVC}  & 21.6 / 52.0  & 90.0 / 43.8 & & 24.8 / 44.6 & 76.3 / 13.4  \\
      MViT~\cite{Xiao2021CVPR} & 57.2 / 73.1 & 85.8 / 24.9  & & \underline{47.8} / \underline{63.7}  & 59.8 / \textbf{\hphantom 06.2} \\
      MixT~\cite{Xie2021NeurIPS}   & \underline{65.7} / \underline{78.1} & \underline{24.6} / \underline{12.4} & & 40.3 / 51.3 & \underline{23.0} / 17.1 \\
      {\small Swin-B}~\cite{Liu2021ICCV} & \textbf{ 78.5} / \textbf{85.4} & \textbf{11.8} / \textbf{\hphantom 06.9}  & & \textbf{61.0} / \textbf{70.8} & \textbf{10.6} / \hphantom 0\underline{6.3} \\

      \bottomrule
    \end{tabular}
    \caption{\textbf{Ablation study on the backbone.}
    We show the effect of varying the backbone used for feature extraction on the OoD performance. Comparing the results before and after fine-tuning with the proposed method, we observe clear improvements in the performance in all backbones.
    }
    \label{tab:backbones_v3}
\end{table}

\boldparagraph{Different Backbones} We use the same Mask2Former model with Swin-B backbone~\cite{Liu2021ICCV} in all our experiments. %
In \tabref{tab:backbones_v3}, we report the results with different backbones including transformer-based Multiscale ViT (MViT)~\cite{Xiao2021CVPR} and Mix Transformer (MixT)~\cite{Xie2021NeurIPS} as well as convolutional WideResnet38 (WR38)~\cite{Zagoruyko2016BMVC} and ResNet101 (R101)~\cite{He2016CVPR} backbones. We keep all the other parameters the same as the default version for a fair comparison.
Fine-tuning with RbA brings consistent improvements in all metrics for all backbones. While Swin-B performs the best, R101, MViT, and MixT can still outperform previous methods on Road Anomaly and achieve competitive results on Fishyscapes LaF.
This experiment shows that the proposed scoring function improves the performance regardless of the backbone.

\section{Conclusion and Future Work}
\label{sec:conclusion}
In this work, we explore the potential of mask classification to segment unknown classes. 
We show that object queries behave like one \vs all classifiers and their independent behavior reduces the overconfidence issue in the predicted scores, resulting in improvements in the performance of the existing scoring-based methods such as max logit and energy-based methods. By treating the result of mask classification as multiple one \vs all classifiers, we propose a novel outlier scoring function called RbA defined in terms of known class probabilities. We also propose an objective to optimize the RbA with limited outlier data, obtaining significant performance gains without affecting the closed-set performance. We show that the RbA eliminates irrelevant sources of uncertainty, such as inlier boundaries and ambiguous background regions, leading to a considerable decrease in false positive rates. Moreover, our proposed method can preserve objectness and smoothness due to the region-level inductive biases learned by the mask classifier.

As this work represents an initial attempt to utilize mask classification for unknown segmentation, its properties can be further explored with potential improvements. Given the increased ability to preserve objectness, open-world incremental learning is one step closer, as unknown masks are more reliable as a source of supervision. While current efforts are limited to static image datasets, temporal or depth information can provide important cues to detect unknowns.

\boldparagraph{Acknowledgements}
We thank A. Kaan Akan, Ali Safaya, G\"orkay Aydemir, Shadi Hamdan, Mert Çökelek, and Merve Rabia Barın for their valuable feedback. This project is funded by the Royal Society Newton Fund Advanced Fellowship (NAF\textbackslash R\textbackslash 2202237). Nazir Nayal and Misra Yavuz are supported by the KUIS AI fellowship. We also thank Unvest R\&D Center for their support.

{\small
\bibliographystyle{ieee_fullname}
\bibliography{bibliography_short, sec/11_references}
}

\clearpage \noindent{\Large \textbf{Appendix}} \appendix
\label{sec:appendix}
\section*{Overview}
This supplementary document contains the implementation details that are necessary to reproduce our approach (\secref{sec:imp_details}), additional ablation results to justify hyper-parameter choices (\secref{sec:add_quantitative}), additional details about analysis and ablation experiments reported in the main paper (\secref{sec:details}), additional qualitative results to showcase our method compared to state-of-the-art (\secref{sec:add_qual}), and some challenging cases that cause failure (\secref{sec:failure}). %

\section{Implementation Details}
\label{sec:imp_details}
\subsection{Architecture} 
\begin{figure*}[t!]
\centering
    \includegraphics[width=\linewidth] {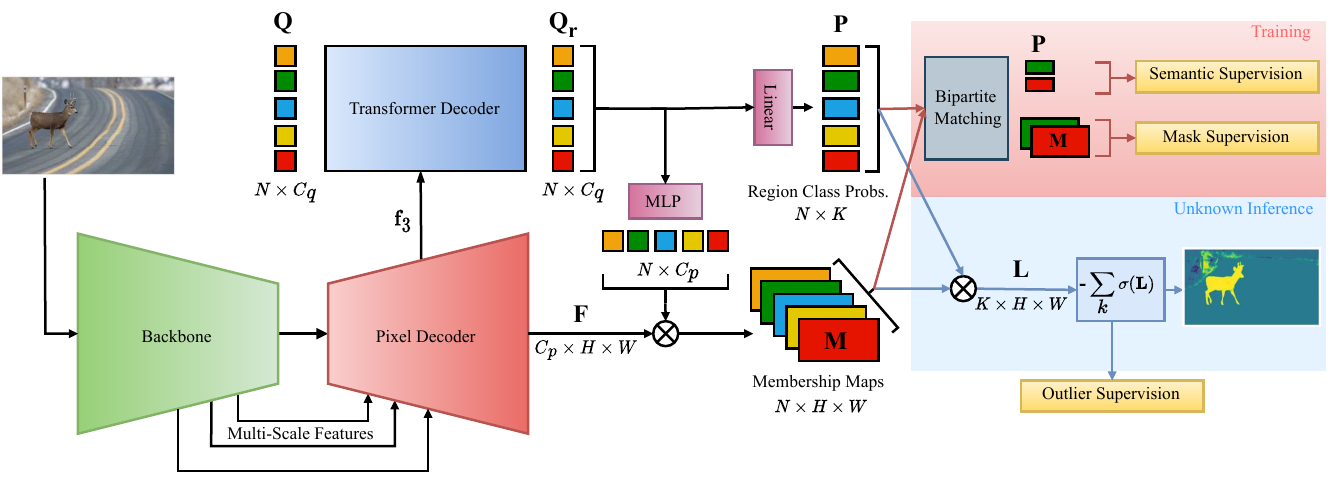}
    \vspace{-0.1in}
    \caption{\textbf{Detailed architecture.} This figure provides a more detailed view of the Mask2Former \cite{Cheng2022CVPR} architecture, including our modifications and unknown inference computation. We use a single transformer decoder layer as opposed to the original implementation that uses 9 layers. Therefore, only a single scale feature $\mathbf{f}_3$ from the last layer is passed from the pixel decoder to the transformer decoder. For outlier supervision, all modules are frozen except for the MLP and the linear layers shown in pink.}
    \vspace{-.1in}
    \label{fig:full_mask2former}
\end{figure*}

\figref{fig:full_mask2former} illustrates the full Mask2Former architecture along with our unknown inference procedure. The main components of the model are the backbone, the pixel decoder, and the transformer decoder. We explain the details of each next.

\boldparagraph{Backbone} We use the Swin-B variant as the backbone~\cite{Liu2021ICCV}. It can take an RGB image with any resolution higher than $32 \times 32$ as input and outputs feature maps at several resolutions to the pixel decoder. Specifically, the output feature maps are downsampled with strides 4~($\bx_4$), 8~($\bx_8$), 16~($\bx_{16}$), and 32~($\bx_{32}$) with respect to the input image.

\boldparagraph{Pixel Decoder} Following~\cite{Cheng2022CVPR}, the pixel decoder mainly consists of 6 layers of deformable attention (MSDeformAttn)~\cite{Zhu2021ICLR}. The multi-scale feature maps with strides $\bx_8$ $\bx_{16}$, and $\bx_{32}$ are processed with MSDeformAttn layers to produce $\mathbf{f}_1$, $\mathbf{f}_2$, and $\mathbf{f}_3$, respectively. In \cite{Cheng2022CVPR}, the three processed feature maps are passed to 9 transformer decoder layers in a round-robin fashion. However, we found that using a single layer in the transformer decoder works better for unknown detection. Therefore, we only pass the last layer $\mathbf{f}_{3}$ to the transformer decoder.
The feature map $\bx_{4}$ is processed with a $1 \times 1$ filter-size convolutional layer and then added to the processed feature map $\mathbf{f}_1$ after bilinear upsampling. Finally, the output is passed to a $3 \times 3$ convolutional layer to produce per-pixel features $\bF \in \mathbb{R}^{C_p \times H \times W}$, where $C_p = 256$ is the embedding dimension. The computation can be summarized as follows:
\begin{equation}
    \bF = \text{Conv}_{3\times3}\left(\text{Conv}_{1\times1}(\bx_{4}) + \text{Upsample}(\mathbf{f}_{1})\right)
\end{equation}

\boldparagraph{Transformer Decoder} Learnable object queries $\bQ \in \mathbb{R}^{N \times C_q}$ are fed to the transformer decoder layer to be processed with the feature maps from the pixel decoder, where $N = 100$ is the number of object queries and $C_q = 256$ is the embedding dimension. A single transformer decoder layer consists of a cross-attention layer followed by self-attention and feed-forward network (FFN), each of which is followed by a LayerNorm. The cross-attention operation is performed with mask-attention, where each object query only attends to regions it predicted in the previous layer. Since we use only a single layer, each object query attends to the region it predicts directly from the input feature map before being processed in the transformer decoder. 
Learnable positional embeddings are added to the object queries. %
The transformer decoder outputs a refined set of object queries $\bQ_r$ that predict the regions and classify them.

\boldparagraph{Region Class Prediction} The refined object queries $\bQ_r$ are fed into a single linear layer followed by a softmax to produce the class probability of each region $\bP \in \mathbb{R}^{N \times K}$ where $K$ is the number of classes.

\boldparagraph{Membership Maps Prediction} The refined object queries $\bQ_r$ are also fed into a 3-layer MLP, so that $\bQ_r$'s dimensionality matches that of $\bF$. Then, $\bQ_r$ and $\bF$ are multiplied before being fed into a sigmoid activation to produce the per-pixel membership maps $\bM \in \mathbb{R}^{N \times H \times W}$.

\subsection{Closed-Set Training}
\boldparagraph{Loss Functions} Before applying any loss function, bipartite matching is used to match object queries to ground truth binary masks, where each mask contains all the pixels of a certain class. The matching cost is computed as a weighted sum of the individual losses. The classification is performed with the cross-entropy loss. A weighted combination of dice loss and binary cross-entropy is used to predict regions.

\boldparagraph{Hyper-Parameters} Following \cite{Cheng2022CVPR}, the model is trained for 90K iterations using a batch size of 16. AdamW~\cite{Loshchilov2019ICLR} optimizer is used with $0.05$ for weight decay and an initial learning rate of $10^{-4}$, which is reduced using a polynomial scheduler. The learning rate for the backbone is multiplied by $0.1$.

\boldparagraph{Data Augmentation} We use the same augmentations as in \cite{Cheng2022CVPR}. First, the short side of the input image is resized by a scale uniformly chosen between $[0.5 - 2]$. Then a random crop of size $512 \times 1024$ is applied. After that, large-scale jittering augmentation \cite{Du2021ARXIV, Ghiasi2021CVPR} is applied with a random horizontal flip.

\subsection{Outlier Supervision}
\boldparagraph{Data Sampling} We use a slightly modified version of AnomalyMix proposed in~\cite{Tian2022ECCV} for outlier supervision. After eliminating the samples that contain Cityscapes classes~\cite{Cordts2016CVPR}, around 40K images remain for outlier supervision on the COCO~\cite{Lin2014ECCV} dataset. For a single fine-tuning experiment, we randomly sample 300 images and fix them throughout the entire fine-tuning phase.

\boldparagraph{Fine-tuned Components} For all the fine-tuning experiments, we only fine-tune the 3-layer MLP and linear layers shown in pink in \figref{fig:full_mask2former}. Their weights together constitute approximately $0.21\%$ of the entire model parameters. 

\boldparagraph{Hyper-Parameters} After the model is trained on the closed-set setting, we fine-tune it for 2000 iterations on Cityscapes~\cite{Cordts2016CVPR} using the setting of the closed-set training; AdamW~\cite{Loshchilov2019ICLR} optimizer with $0.05$ weight decay and $10^{-4}$ initial learning with polynomial scheduling. For every Cityscapes image used in fine-tuning, an object from the 300 COCO samples is uniformly chosen and pasted on the Cityscapes image with probability $p_{out} = 0.1$, which is independent for each image. The 
RbA score for outlier pixels is optimized with the squared hinge loss $\mathcal{L}_{RbA}$ using $\alpha = 5$.

\section{Additional Ablation Study}
\label{sec:add_quantitative}

\begin{figure*}[t!]
    \centering
    \begin{subfigure}{0.3\linewidth}
        \includegraphics[width=\linewidth]{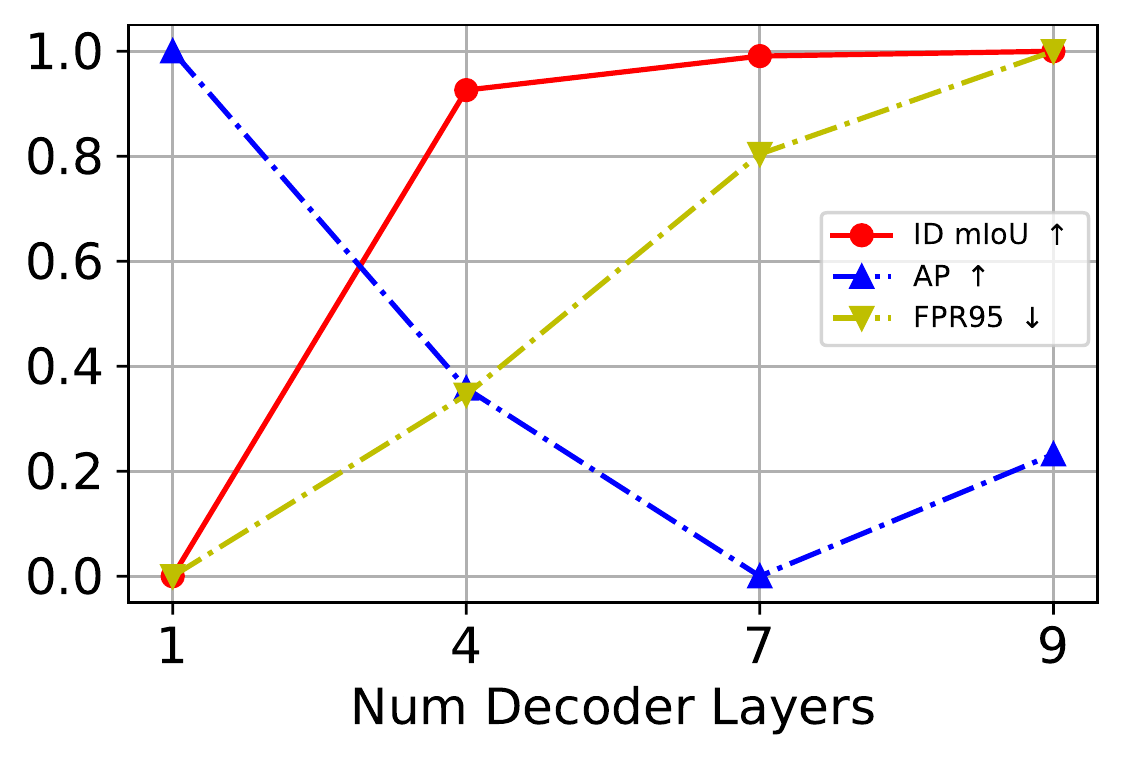}
        \caption{Number of Decoder Layers}
        \label{fig:dec_layer_metrics}
    \end{subfigure}
    \hfill
    \begin{subfigure}{0.3\linewidth}
    \centering
        \includegraphics[width=\linewidth]{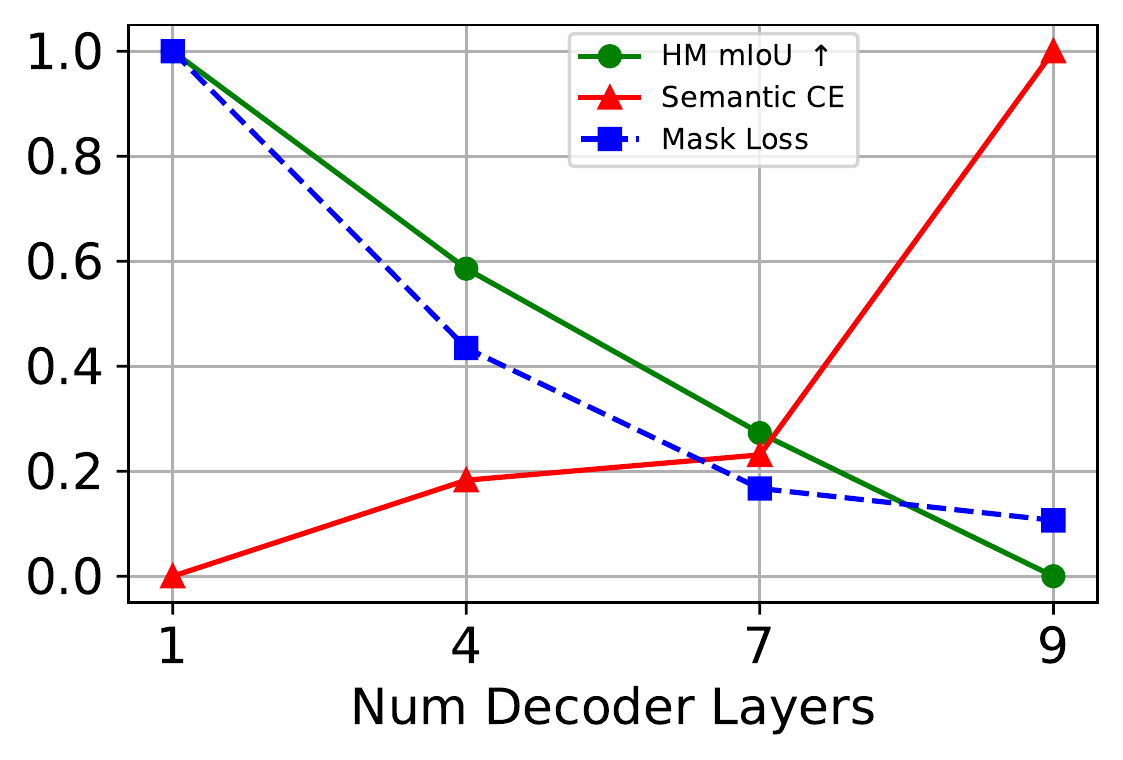}
        \caption{Behavior of Different Losses}
        \label{fig:dec_layer_val_loss}
    \end{subfigure}
    \hfill
    \begin{subfigure}{0.3\linewidth}
    \includegraphics[width=\linewidth] {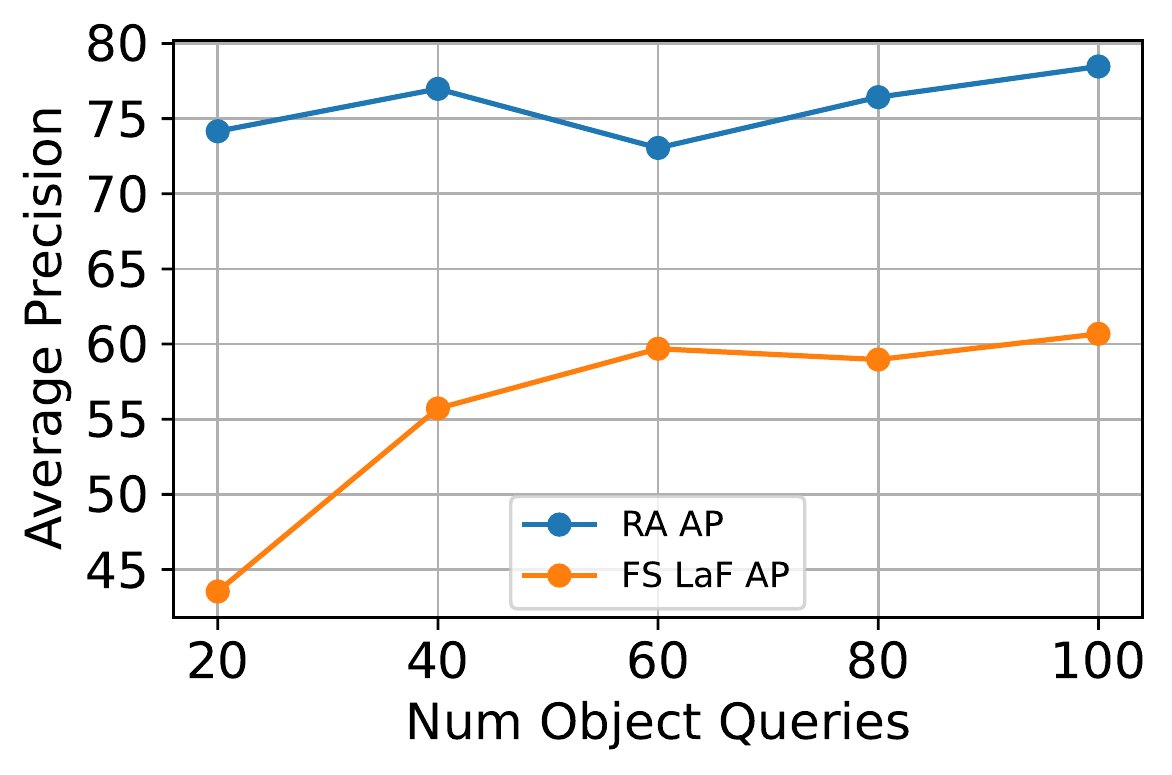}
    \caption{Number of Object Queries}
    \label{fig:object_queries_ablation}
    \end{subfigure}
    \caption{\textbf{Ablation study on architectural choices.} 
    With more decoder layers, in-distribution performance on Cityscapes improves but the outlier performance drops in terms of AP and FPR@95 on Road Anomaly \subref{fig:dec_layer_metrics}.
    Good performance is mainly due to better mask prediction at the cost of a higher semantics loss \subref{fig:dec_layer_val_loss}. The drop in mIoU with hard masking (\textbf{HM mIoU}) is another indicator of semantic information loss in the object queries with more decoder layers. The performance improves as the number of object queries increases \subref{fig:object_queries_ablation}.}
\end{figure*}

\begin{table*}[t!]
\footnotesize
\begin{subtable}{0.28\linewidth}
    \centering
    \begin{tabular}{c c c} 
    \toprule
    Num Iter        & AP $\uparrow$ & FPR@95 $\downarrow$ \\
    \midrule
    1000              & 83.20 $\pm$ 0.11 & 8.69 $\pm$ 0.04 \\
    2000              & \underline{85.49} $\pm$ 0.08 & \textbf{7.25} $\pm$ 0.04 \\
    3000             & \textbf{85.72} $\pm$ 0.12 & 7.82 $\pm$ 0.11 \\
    4000              & 84.89 $\pm$ 0.07 & \underline{7.45} $\pm$ 0.05 \\
    5000              & 84.66 $\pm$ 0.06 & 8.14 $\pm$ 0.03 \\
    \bottomrule
    \end{tabular}
    \caption{Number of Iterations} 
    \label{tab:mlp_abl_iter}       
\end{subtable}
\hfill
\begin{subtable}{.34\linewidth}
    \centering
    \begin{tabular}{c c c c} 
    \toprule
    $p_{out}$       & AP $\uparrow$ & FPR@95 $\downarrow$ & mIoU $\uparrow$   \\
    \midrule
    0.05            & 84.34 $\pm$ 0.15 & 9.12 $\pm$ 0.10 & \textbf{82.28} $\pm$ 0.02 \\
    0.1             & \underline{85.48} $\pm$ 0.12 & \textbf{7.24} $\pm$ 0.06 & \underline{82.15} $\pm$ 0.09 \\
    0.2             & \textbf{85.74} $\pm$ 0.11 & \underline{7.90} $\pm$ 0.11 & 81.54 $\pm$ 0.02 \\
    0.4             & 84.97 $\pm$ 0.11 & 9.19 $\pm$ 0.05 & 81.27 $\pm$ 0.06\\    
    \bottomrule
    \end{tabular}
    \caption{Outlier Selection Probability} 
    \label{tab:mlp_p_out_ablation}
\end{subtable}
\hfill
\begin{subtable}{.28\linewidth}
    \centering
    \begin{tabular}{c  c  c} 
    \toprule
    $\alpha$ & AP $\uparrow$ & FPR@95 $\downarrow$ \\
    \midrule
    -0.1        & 84.01 $\pm$ 0.18 & 9.25 $\pm$ 0.06 \\
    -0.01       & 84.41 $\pm$ 0.14 & 9.06 $\pm$ 0.11 \\
    0.0         & 84.30 $\pm$ 0.06 & 9.10 $\pm$ 0.07 \\
    2           & \underline{85.30} $\pm$ 0.07 & 7.80 $\pm$ 0.06 \\
    5           & 85.24 $\pm$ 0.08 & \textbf{6.95} $\pm$ 0.08 \\
    10          & \textbf{85.61} $\pm$ 0.10 & \underline{7.26} $\pm$ 0.07 \\
    \bottomrule
    \end{tabular}
    \caption{Outlier Threshold} 
    \label{tab:mlp_alpha_out_ablation}
\end{subtable}
\caption{\textbf{Ablation study on outlier supervision.} Finetuning with RbA loss for 2000-3000 iterations achieves the best performance and the performance deteriorates after 3000 iterations as shown in \subref{tab:mlp_abl_iter}. A higher probability of exposure to outlier data results in a consistent decline in the closed-set performance. A probability of 0.1 achieves the best balance between outlier and inlier performance \subref{tab:mlp_p_out_ablation}. Finally, we ablate the RbA loss parameter $\alpha$ in \subref{tab:mlp_alpha_out_ablation} and find that the best results are achieved with $\alpha > 0$.
}
\end{table*}

\begin{table}{}
    \centering
    \adjustbox{max width=\columnwidth}{%
    \begin{tabular}{c c c c c c c} 
    \toprule
    \multirow{2}{*}{Module} & \multirow{2}{*}{Params (\%)}  & \multirow{2}{*}{mIoU} & \multicolumn{2}{c}{Road Anomaly} & \multicolumn{2}{c}{FS LaF} \\ 
        \cmidrule(r){4-5} \cmidrule(l){6-7}
         & & & AP $\uparrow$ & FPR $\downarrow$ & AP $\uparrow$ & FPR $\downarrow$ \\
    \midrule
    Full Model          & 100 & 80.81 & 76.00 & \underline{9.50} & \textbf{73.88} & \underline{6.02}  \\
    Transformer Dec.    & 1.93 & 80.24 &\underline{ 85.08} & 10.18 & \underline{72.6} & \textbf{5.51} \\
    Pixel Dec.          & 4.66 & \underline{81.59} & 75.83	& 10.51 & 69.8 & 6.77  \\
    MLP+Linear          & \textbf{0.21} & \textbf{82.20} & \textbf{85.42} & \textbf{6.92} & 70.81 & 6.30  \\
    \bottomrule
    \end{tabular}}
    \caption{\textbf{Ablation study on fine-tuning different modules.} We show the effect of fine-tuning different components of the model on the Road Anomaly and Fishyscapes LaF validation sets. Fine-tuning MLP+Linear maintains the best performance in unknown detection without sacrificing the closed-set performance.} 
    \label{tab:supp_ablation}
\end{table}

\boldparagraph{Number of Transformer Decoder Layers} As shown in \cite{Cheng2022CVPR}, more transformer decoder layers improve the inlier performance, \ie mIoU on Cityscapes. However, we found that using fewer decoder layers results in better performance in terms of outliers. \figref{fig:dec_layer_metrics} highlights the decrease in performance in terms of the AP and FPR@95 on the Road Anomaly dataset as the number of decoder layers increases. We investigate this behavior by isolating the sources of error with respect to the number of decoder layers. \figref{fig:dec_layer_val_loss} shows semantic and mask losses of the Mask2Former~\cite{Cheng2022CVPR} averaged over the validation samples on Cityscapes. With more decoder layers, we observe that the semantic cross-entropy loss increases while the mask-related BCE and dice losses decrease. This shows that the increase in inlier mIoU with more decoder layers can be attributed to increased performance in detecting masks at the cost of a higher semantic error. By using fewer decoder layers, we regulate the semantic confusion, which helps to better align the logit scores, resulting in better outlier performance. 

\figref{fig:dec_layer_val_loss} also shows the mIoU evaluated by applying hard masking on the specialized object queries. Specialized object queries perform worse with more decoder layers. The information loss in the object queries as well as the increase in the semantic loss show the importance of semantics for outlier segmentation compared to precise masks.

\boldparagraph{Number of Object Queries} The original Mask2Former \cite{Cheng2022CVPR} uses 100 object queries. We train different models by varying the number of object queries to observe its effect on detecting outliers. We focus on the AP values on both Road Anomaly~\cite{Lis2019ICCV} with large objects and on Fishyscapes LaF~\cite{Blum2021IJCV} with small objects. \figref{fig:object_queries_ablation} shows that more object queries result in better AP on both datasets. Even if some object queries specialize in predicting a particular class, the other object queries still play a role, especially for rare classes, as shown in Fig.3 in the main paper.

\boldparagraph{Outlier Data Exposure} We perform an experiment to show the effect of the number of iterations required for fine-tuning with our RbA loss in \tabref{tab:mlp_abl_iter}. We report the AP and FPR metrics on the Road Anomaly dataset, averaged over 5 different runs to eliminate the effect of randomness. We can see that the best results can be achieved after around 2000 and 3000 iterations and then begin to degrade. We also evaluate the effect of the amount of outlier data exposed during training, which is controlled by the parameter $p_{out}$. We experiment with different values for $p_{out}$ and report the outlier performance on Road Anomaly and closed-set performance on Cityscapes averaged over 5 different runs in \tabref{tab:mlp_p_out_ablation}. We can see that more exposure to outlier data negatively affects the closed-set performance. Consequently, even the outlier segmentation performance starts to degrade for $p_{out} > 0.2$. We choose the $p_{out} = 0.1$ because it strikes a reasonable balance between outlier and closed-set performance.

\boldparagraph{RbA Loss Parameter} In \tabref{tab:mlp_alpha_out_ablation}, we report the performance of our loss function $ \cL_{\mathrm{RbA}} $ using different values of $\alpha$, averaged over 5 different runs. We find that positive values of $\alpha$ work similarly well and set $\alpha$ to 5 in our experiments. %

\boldparagraph{Fine-tuned Component} In \tabref{tab:supp_ablation}, we analyze the effect of fine-tuning different parts of the model on validation sets of Road Anomaly~\cite{Lis2019ICCV} and Fishyscapes Lost and Found~\cite{Blum2021IJCV} using RbA loss. The alternative components we experimented with are the full model, only the transformer decoder (blue + pink in \figref{fig:full_mask2former}), only the pixel decoder (red in \figref{fig:full_mask2former}), and MLP+Linear layers (pink in \figref{fig:full_mask2former}). On the Road Anomaly dataset, fine-tuning MLP+Linear achieves the best performance in terms of AP and FPR.
On Fishyscapes LaF, the best AP is achieved by fine-tuning the entire model, while the best FPR is obtained by fine-tuning only the transformer decoder. Both options cost a decrease in performance on Road Anomaly and negatively affect the closed-set performance. Fine-tuning MLP+Linear achieves the best balance between outlier detection and closed-set performance and is the least costly option in terms of the number of parameters finetuned.

\section{Details of Experiments}
\label{sec:details}
In this section, we provide further illustrations and detailed settings of our analysis and ablation experiments in the main paper. We first perform an experiment to verify the specialization of object queries. We then provide a detailed formulation of our ablations on the loss functions and other methods using Mask2Former including the hyper-parameters that we use to obtain the results presented in the main paper.

\subsection{Specialization of Object Queries}
\begin{figure*}[t!]
    \centering
    \includegraphics[width=\linewidth] {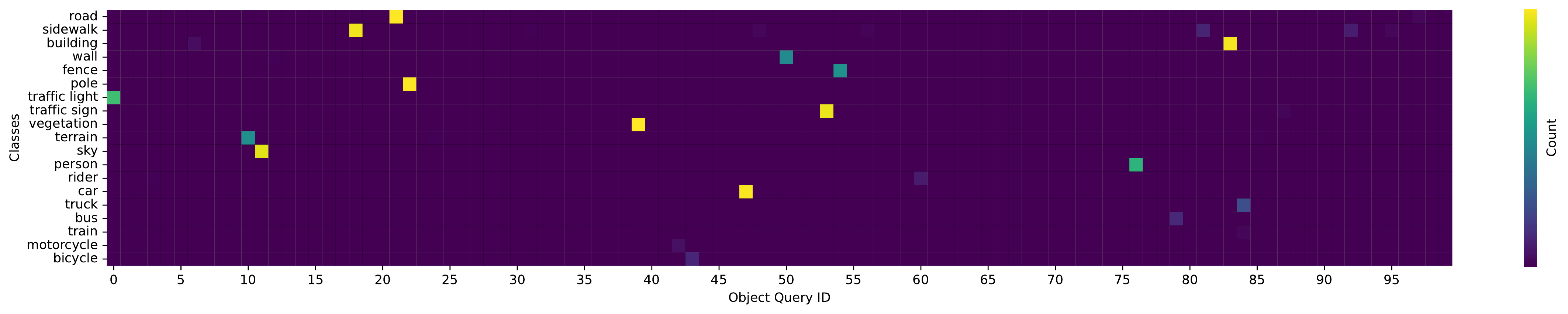}
    \label{fig:oq_vote}
    \vspace{-0.2in}
    \caption{\textbf{Specialization of Object Queries.} Certain object queries specialize in predicting a specific class. For each query, we show how many times it predicts a region to belong to a class with high confidence. The sparsity in the plot clearly shows the specialization of queries.}
    \label{fig:one_vs_all}
\end{figure*}

Our method is based on our finding that the object queries in mask classification enjoy a degree of independence from one another and that each object query in a subset specializes in segmenting a specific class from the closed set. Due to bipartite matching being applied between queries and ground truth class masks during training, this behavior can be anticipated. Here, we empirically verify it using a different dataset than the one used in training (BDD100K \cite{Yu2020CVPR}). For each object query, we count how many times it predicts a certain class with high confidence. \figref{fig:one_vs_all} shows the  heatmap of counts for the Mask2Former model with 100 object queries. For each of the closed-set classes, there is a single object query dominantly predicting it. 

\begin{figure}
    \centering
    \includegraphics[width=.5\linewidth]{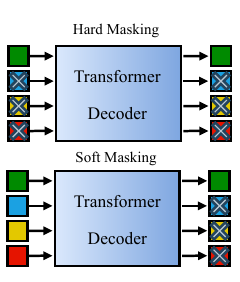}
    \caption{\textbf{Illustration of hard vs. soft masking of object queries.} In hard masking, when predicting class $k$, all but the object queries specialized to predict class $k$ are masked before the transformer decoder so that the specialized query can only interact with the image features. In soft masking, the queries are allowed to interact within the transformer decoder but are dropped after, and only the specialized query is used to predict class $k$.}
    \label{fig:hard_soft_masking}
\end{figure}

In the main paper, we test the independence of the specialized queries by applying hard masking and soft masking and evaluating per class IoU on Cityscapes. \figref{fig:hard_soft_masking} shows an illustration of hard and soft masking applied.

\subsection{Other Loss Functions}
In the main paper, we perform an ablation study with different loss functions in comparison to squared hinge loss. In this section, we provide the formulation and the parameter setting of each loss function. In the following, $\Omega_{out}$ denotes the set of outlier pixels, and $\Omega_{in}$, the set of inlier pixels on an image.

\boldparagraph{Mean-Squared Error (MSE)} With MSE loss, we optimize the RbA to be closer to $\alpha = 5$ for outlier pixels:
\begin{equation}
    \cL_{\mathrm{MSE}} = \sum_{\bx \in \Omega_{out}} \left(\mathrm{RbA}(\bx) - \alpha \right)^2
\end{equation}

\boldparagraph{L1} Similar to MSE, we optimize the RbA with L1 loss by setting $\alpha$ to 5:
\begin{equation}
    \cL_{L1} = \sum_{\bx \in \Omega_{out}} \left\vert \mathrm{RbA}(\bx) - \alpha \right\vert
\end{equation}

\boldparagraph{Binary Cross Entropy (BCE)} We formulate the scoring of outliers as a per-pixel binary classification problem, where outliers correspond to the positive class. We use the RbA score as the logit score for the positive class. Assuming that $y$ corresponds to the binary label (outlier \vs inlier) of a pixel $\bx$, we optimize the BCE loss as follows:
\begin{equation}
    \cL_{\mathrm{BCE}} = \sum_{\bx \in \Omega_{out}} y \cdot \log\left(\mathrm{RbA}(\bx)\right) + (1 - y) \cdot \log\left(1 - \mathrm{RbA}(\bx)\right)
\end{equation}

\boldparagraph{KL Divergence} We use the KL Divergence to minimize the distance between the predicted class distribution to a fixed distribution. For inlier pixels, we minimize the distance to the Dirac delta function where the correct class has a probability of $1.0$. For outlier pixels, we minimize the distance to a uniform distribution where the entropy is maximum. Even though this loss function does not optimize our proposed score function RbA, it helps us estimate the contribution of RbA by pushing the predicted class probabilities toward the ideal distributions for outliers and inliers. Let $\bL_p(\bx)$ be the class probability distribution of a pixel $\bx$ with ground truth label $y$. Let $\bP_{in}(y)$ be a fixed probability distribution where class $y$ has probability 1.0. Let $\unif$ be a fixed uniform probability distribution. Formally, the loss function is defined as follows: 
\begin{equation}
    \begin{split}
        \cL_{in} &= \sum_{\bx \in \Omega_{in}} \infdiv{\bL_p(\bx)}{\bP_{in}(y)} \\
        \cL_{out} &= \sum_{\bx \in \Omega_{out}} \infdiv{\bL_p(\bx)}{\unif} \\
        \cL_{KL} &= \frac{1}{2} \left(\mathcal{L}_{in} + \cL_{out} \right)
    \end{split}
\end{equation}

\subsection{Other Methods with Mask2Former}
To disentangle the contribution of our scoring function RbA from the architecture, in the main paper, we present the results of the state-of-the-art methods PEBAL \cite{Tian2022ECCV} and DenseHybrid \cite{Grcic2022ECCV} using the Mask2Former architecture. Here, we provide the details of this ablation experiment for each method. For a fair comparison, we train both methods by fine-tuning the same components as the RbA, that is the MLP+Linear layers as shown in \figref{fig:full_mask2former}, and also using the same outlier data supervision method as described in the main paper.

\boldparagraph{PEBAL} The optimized objective as per the official implementation \cite{Tian2022ECCV} consists of three components. First, there is the pixel-wise anomaly abstention loss (PAL) with the abstention term and penalty defined as follows:
\begin{equation*}
    \mathcal{L}_{PAL} = -\sum_{\bx \in \Omega} \log \left( \bL_y(\bx) + \frac{\bL_{K+1}(\bx)}{a(\bx)}\right)
\end{equation*}
where $\Omega$ is the set of all pixels on a given image, $y \in {1, \dots, K+1}$ is the ground truth class of pixel $\bx \in \Omega$, and $K+1$ is the class for the outliers. In Mask2Former, class $K+1$ is assumed to be the no object class, therefore we avoid dropping it from the region class probabilities term $\bP$ (see \figref{fig:full_mask2former}) while fine-tuning with the PEBAL objective. The abstention penalty term $a(\bx)$ is defined as follows:
\begin{equation}
    \begin{split}
        a(\bx) &= \left( -E(\bx)\right)^2 \\
        E(\bx) &= - \log \sum_{k=1}^{K} \exp(\bL_k(\bx))
    \end{split}
\end{equation}
where $E(\bx)$ is the free energy function. When the penalty term is high, the prediction is discouraged from abstaining and vice versa. 

The second component of the loss optimizes the energy terms such that it is maximized for outlier pixels and minimized for inlier pixels as follows:
\begin{equation}
    \begin{split}
        \mathcal{L}_{energy} =& \sum_{\bx \in \Omega_{in}} \max(0, E(\bx) - m_{in})^2 + \\
        & \sum_{\bx \in \Omega_{out}} \max(0, m_{out} - E(\bx))^2
    \end{split}
\end{equation}
where $m_{in}$ and $m_{out}$ are hyper-parameters to be set. In our experiments, we experimentally use $m_{out} = -2.5$ and $m_{in} = -3.5$. 

The last component is a regularization term for the smoothness and sparsity of the predicted energy map:
\begin{equation}
    \mathcal{L}_{reg} = \sum_{\bx \in \Omega} \beta_1 \vert E(\bx) - E(\mathcal{N}(\bx))\vert + \beta_2 \vert E(\bx) \vert
\end{equation}
where $\mathcal{N}(\bx)$ is the set of vertical and horizontal neighboring pixels of $\bx$, and $\beta_1$ and $\beta_2$ are hyper-parameters. We use $\beta_1 = 3 \times 10^{-7}$ and $\beta_2 = 5 \times 10^{-5}$. 

The full objective is defined as the weighted sum of the three loss functions:
\begin{equation}
    \cL_{PEBAL} = \cL_{PAL} + \beta \cL_{energy} + \cL_{reg}
\end{equation}
where we set $\beta = 0.1$. Starting from the same checkpoint that we use fine-tuning RbA, we optimize the model with $\cL_{PEBAL}$ for 5K iterations. We set other hyper-parameters to be the same as the ones that we use for RbA.

\boldparagraph{DenseHybrid} Following the official implementation of DenseHybrid~\cite{Grcic2022ECCV}, we added an additional outlier prediction head $\bD(\bx) \in \mathbb{R}^{2 \times H \times W}$ to the Mask2Former model which is defined as follows:
\begin{equation}
    \bD(\bx) = \mathrm{Conv}_{3 \times 3}\left(\mathrm{ReLU}(\mathrm{BatchNorm}(\bx))\right)
\end{equation}
The outlier prediction head takes the output feature map of the highest resolution from the pixel decoder and predicts a binary output for every pixel denoting the probability of being an outlier. The objective for DenseHybrid is defined as follows:
\begin{equation}
    \cL_{DH} = \mathrm{CE}(\bL(\bx_{in}), \bY_{in}) + \beta_1 \mathrm{CE}(D(\bx), \bY_{out}) + \beta_2 \mathcal{L}_{o}
\end{equation}
where $\mathrm{CE}$ is short for the cross-entropy loss, $\bx_{in} \in \Omega_{in}$ denotes the set of inlier pixels, $\bY_{in}$ denotes the ground truth for the closed-set and $\bY_{out}$ denotes the binary map where the outlier pixels are set to one. We experimentally set the hyper-parameters $\beta_1=0.3$ and $\beta_2=0.03$. $\cL_o$ is defined as follows:
\begin{equation}
    \cL_o = \frac{1}{\vert \Omega_{out} \vert}\sum_{\bx \in \Omega_{out}} \log \sum_{k=1}^{K} \exp \left(\bL_k(\bx)\right) + \mathrm{sg}[\mathrm{mean}(\bL(\bx))]
\end{equation}
where $\mathrm{sg}$ is short for the stop gradient operation, and $\mathrm{mean}$ denotes the mean of all the elements of the input tensor.

\section{Additional Qualitative Results}
\label{sec:add_qual}
\begin{figure*}[t!]
    \centering
    \includegraphics[width=\linewidth] {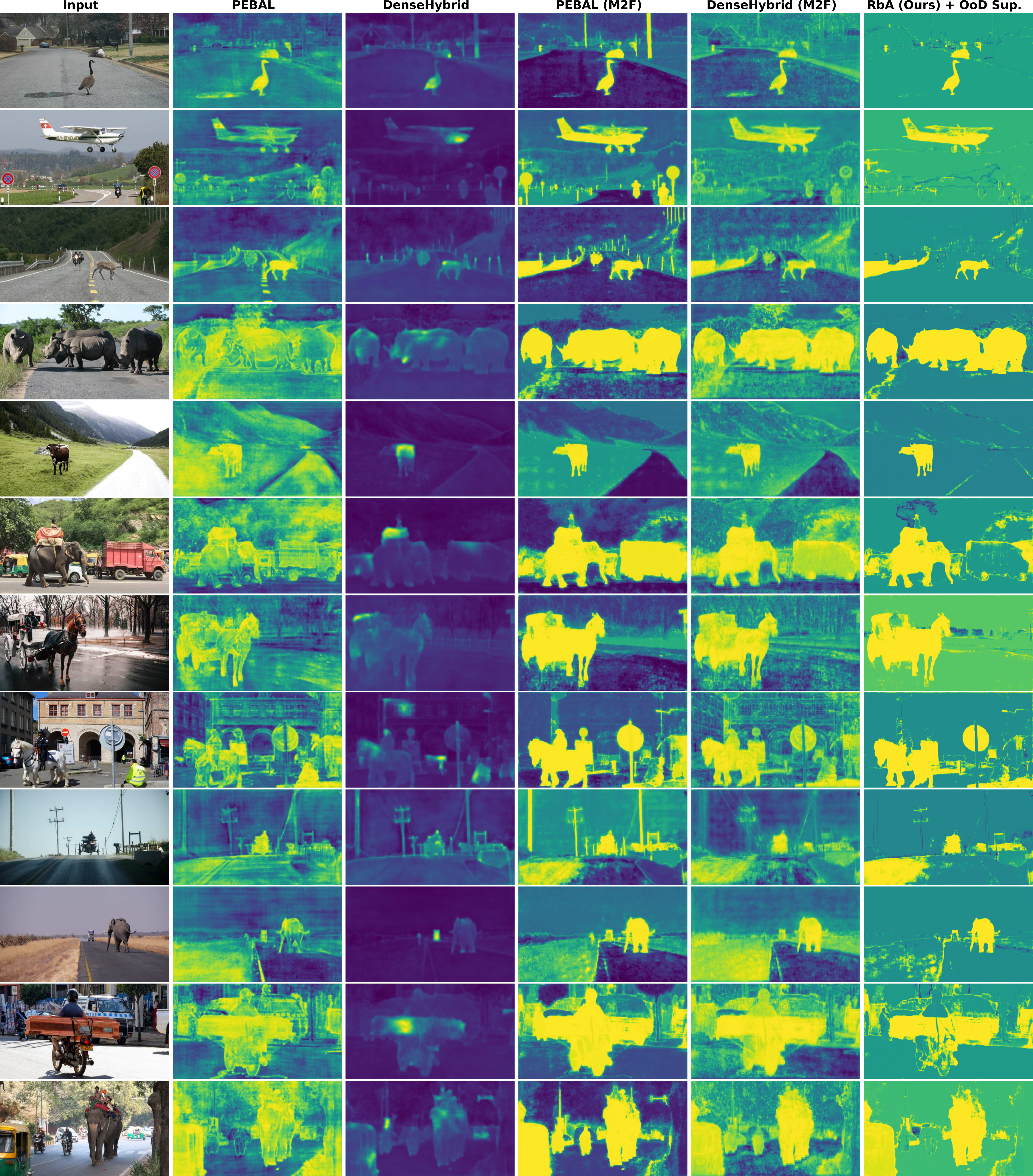}
    \vspace{-0.1in}
    \caption{\textbf{Qualitative Results on SMIYC Anomaly Track.} On the anomaly track of the SMIYC benchmark, we compare RbA with outlier (OoD) supervision to the state-of-the-art methods PEBAL \cite{Tian2022ECCV} and DenseHybrid \cite{Grcic2022ECCV} using the models that were shared in their respective repositories, as well as the versions we trained using Mask2Former (M2F). RbA better distinguishes outliers from inliers and produces more smooth outlier maps with fewer false positives. 
    }
    \vspace{-.1in}
    \label{fig:qual_anomaly_track}
\end{figure*}

\begin{figure*}[t!]
    \centering
    \includegraphics[width=\linewidth] {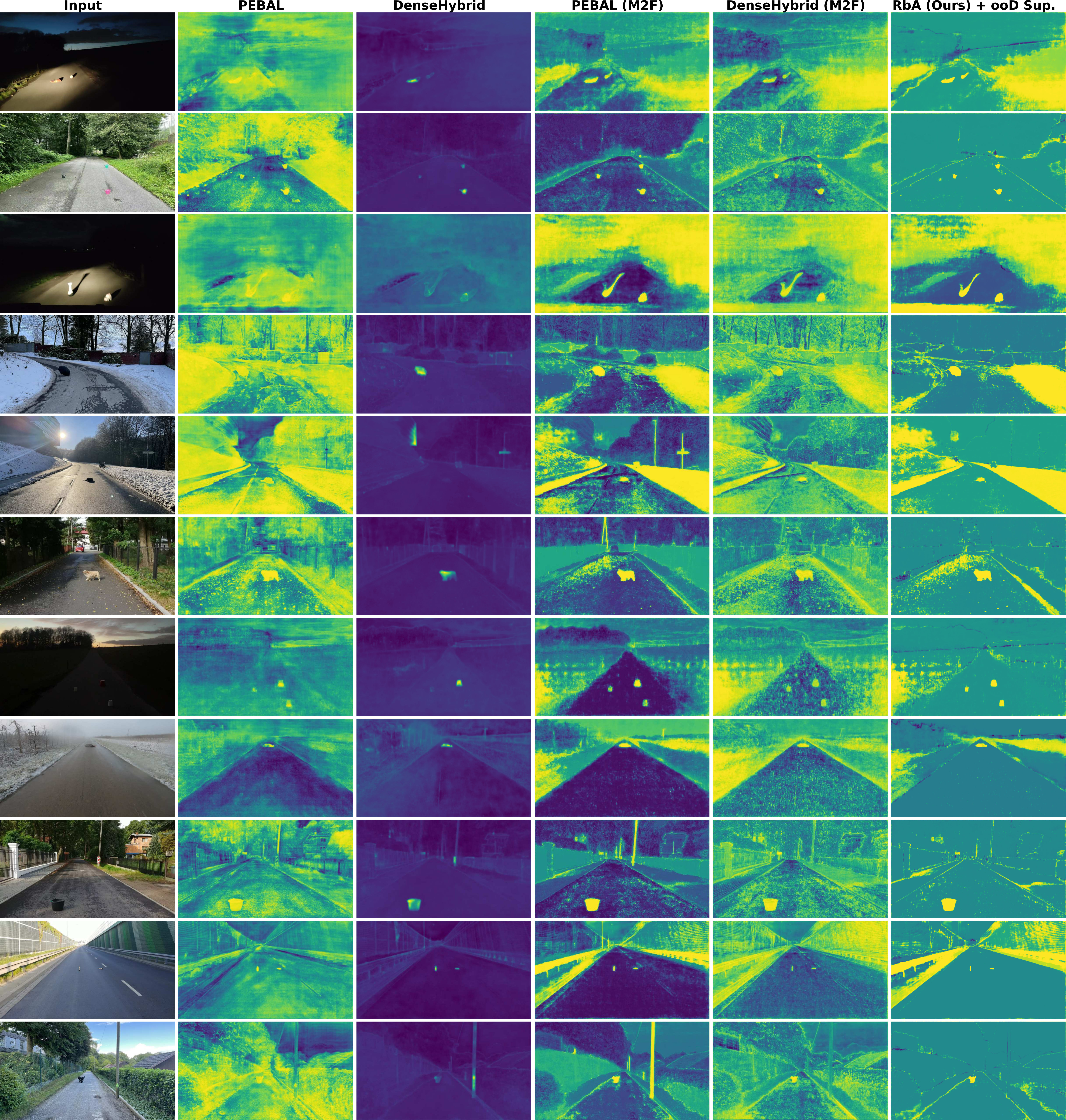}
    \vspace{-0.1in}
    \caption{\textbf{Qualitative Results on SMIYC Obstacle Track.} We compare RbA with outlier supervision to the state-of-the-art methods PEBAL \cite{Tian2022ECCV} and DenseHybrid \cite{Grcic2022ECCV}. Under adverse weather and difficult lighting conditions, RbA can detect anomalies consistently better compared to DenseHybrid and reduce false positives more compared to PEBAL.}
    \vspace{-.1in}
    \label{fig:qual_obstacle_track}
\end{figure*}

In \figref{fig:qual_anomaly_track} and \figref{fig:qual_obstacle_track}, we show additional qualitative results of RbA compared to the state-of-the-art methods PEBAL~\cite{Tian2022ECCV} and DenseHybrid~\cite{Grcic2022ECCV}. For other methods, we show both the outputs of the models reported in their respective repositories and our implementations of the methods using Mask2Former. %
The proposed scoring function RbA reduces the false positives on the boundaries of inliers and ambiguous background regions compared to the baselines. These improvements can be observed more prominently on the obstacle track (\figref{fig:qual_obstacle_track}) under adverse weather and lighting conditions. Moreover, compared to the baselines, RbA results in fewer false positives as a result of reducing confusion with inlier classes.

\section{Failure Cases}
\label{sec:failure}
We analyze some failure cases of our method in this section. A common reason for the failure cases is the high similarity to the inlier classes.

\begin{figure*}[t!]
    \centering
    \includegraphics[width=\linewidth] {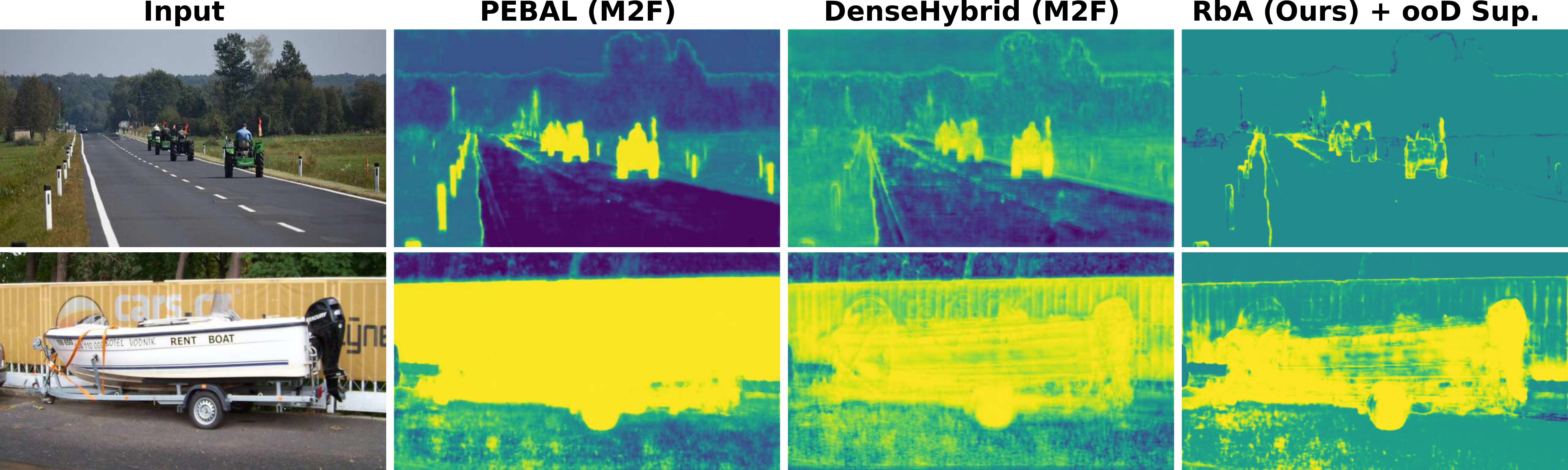}
    \caption{\textbf{Failure Cases: Tractors and Boats.} Due to their high similarity to inlier car and truck classes, unknown objects like tractors or boats are sometimes %
    predicted as inliers.}
    \label{fig:failure_tractor_and_boat}
\end{figure*}

\boldparagraph{Tractors and Boats} As shown in \figref{fig:failure_tractor_and_boat}, %
RbA fails to detect tractors and boats as outliers due to their similarity to inlier vehicle instances. Although the objects are partially identified, the model cannot decisively predict the whole object regions as outliers. The existing methods either segment the boats and tractors at the cost of more false positives, as in the case of PEBAL, or also suffer from a lack of smoothness, as in the case of DenseHybrid.

\begin{figure*}[t!]
    \centering
    \includegraphics[width=\linewidth] {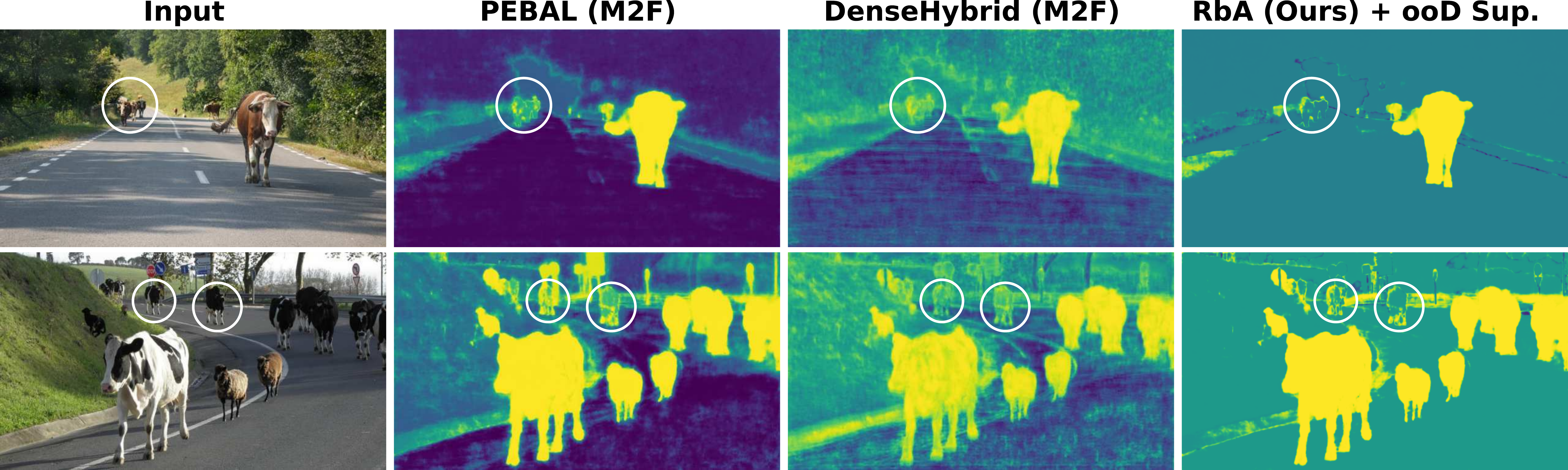}
    \caption{\textbf{Failure Cases: Animals Confused As Pedestrians.} As the pedestrian is one of the most frequent classes on Cityscapes, the model sometimes predicts animals that appear at a distance as pedestrians (highlighted in circles) on images from SMIYC Anomaly Track. } %
    \label{fig:failure_animals}
\end{figure*}

\boldparagraph{Far away Animals} As shown in \figref{fig:failure_animals}, animals that are situated relatively far from the camera are confused as the inlier pedestrian class. This can be attributed to the dominance of pedestrian class in the training data as well as the similarity of legged animals to a pedestrian in appearance.

\begin{figure*}[t!]
    \centering
    \includegraphics[width=\linewidth] {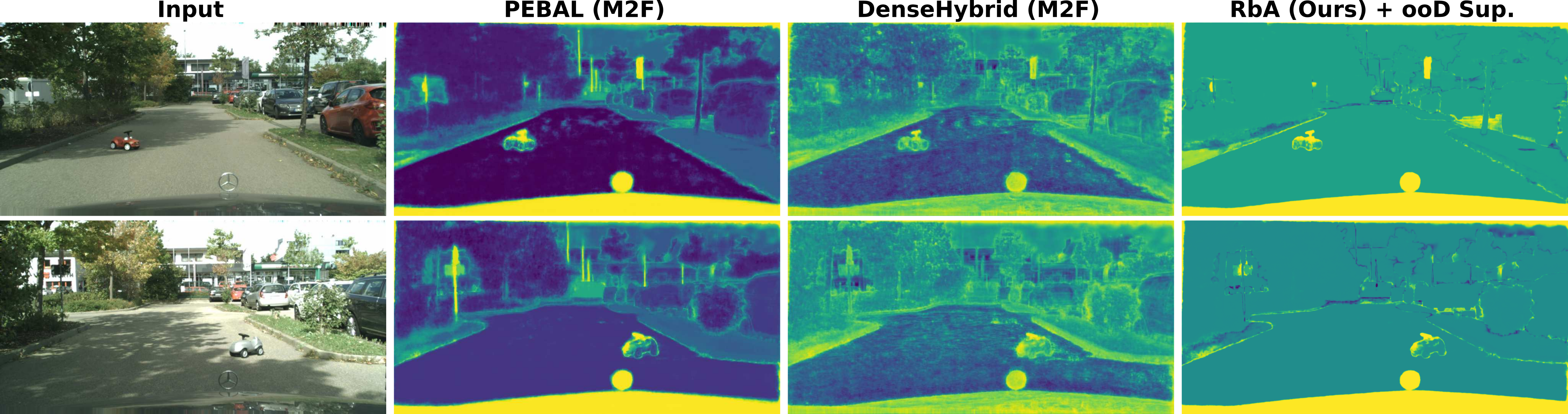}
    \caption{\textbf{Failure Cases: Toy Cars Predicted As Real Cars.} One confusing anomaly case for our model is small toy cars placed in front of the vehicle. Even though they can be semantically considered as cars, they are considered obstacles in a real driving scenario.}
    \label{fig:failure_toy_cars}
\end{figure*}

\boldparagraph{Toy Cars} \figref{fig:failure_toy_cars} shows that the model fails to detect a toy car on the road and predicts it confidently as the inlier car class. While the class assignment can be considered semantically correct, it is still a hazard in a real driving scenario. Note that a small car can either be a toy car or a real car that is far away. Therefore, distinguishing real cars from toy cars might require additional information such as depth or scale.

\clearpage

\clearpage

\clearpage

\clearpage

\end{document}